\documentclass[conference]{IEEEtran}
\IEEEoverridecommandlockouts
\usepackage{cite}
\usepackage{amsmath,amssymb,amsfonts}
\usepackage{graphicx}
\usepackage{textcomp}

\usepackage[utf8]{inputenc} 
\usepackage[T1]{fontenc}    
\usepackage{hyperref}       
\usepackage{url}            
\usepackage{booktabs}       
\usepackage{amsfonts}       
\usepackage{nicefrac}       
\usepackage{microtype}      
\usepackage{lipsum}
\usepackage{fancyhdr}       
\usepackage{graphicx}       
\graphicspath{{media/}}     
\usepackage[dvipsnames]{xcolor}
\usepackage{float}
\usepackage{comment}
\usepackage{wrapfig}
\usepackage{algorithmicx}
\usepackage{algpseudocode}
\usepackage{enumerate}
\usepackage{mathptmx} 
\usepackage[linesnumbered,ruled,vlined]{algorithm2e}

\def\BibTeX{{\rm B\kern-.05em{\sc i\kern-.025em b}\kern-.08em
    T\kern-.1667em\lower.7ex\hbox{E}\kern-.125emX}}
\begin{document}
\sloppy 
\title{An AI-Driven Data Mesh Architecture Enhancing Decision-Making in Infrastructure Construction and Public Procurement*\\
\thanks{This work is the result of a collaborative effort by the extended engineering team at Taiyō.AI, alongside contributions from associated experts and conversations with over 100 key industry leaders who have guided the design and development of this system over the years.}
}

\author{
    \IEEEauthorblockN{Saurabh Mishra\IEEEauthorrefmark{1}, Mahendra Shinde\IEEEauthorrefmark{1}, Aniket Yadav\IEEEauthorrefmark{1}, Bilal Ayyub\IEEEauthorrefmark{2}\IEEEauthorrefmark{1}, Anand Rao\IEEEauthorrefmark{3}\IEEEauthorrefmark{1}}
    \IEEEauthorblockA{\IEEEauthorrefmark{1}Taiyō.AI}
    \IEEEauthorblockA{\IEEEauthorrefmark{2}University of Maryland, College Park}
    \IEEEauthorblockA{\IEEEauthorrefmark{3}Carnegie Mellon University}
}
\maketitle

\begin{abstract}
Infrastructure construction, often referred to as an ``industry of industries'' is deeply intertwined with government spending and public procurement, offering immense potential for enhanced efficiency and productivity through improved transparency and access to foundational information. By capitalizing on this potential, we can achieve significant productivity gains, cost savings, and positive economic impacts across the broader economy. Recognizing this opportunity, we introduce an integrated software ecosystem combining Data Mesh and Service Mesh architecture that encompasses the largest training data for infrastructure, procurement, scientific publications, activities and  risk information with 100B+ tokens along with a systematic AI framework. Our system, underpinned by a Knowledge Graph linked to multi-agents for domain-specific tasks and Q\&A capabilities. By standardizing and ingesting data from diverse sources, we transform raw data into structured knowledge. Leveraging large language models (LLMs) and automation, our system sets new standards for data structuring and automated knowledge creation, assisting in decision-making for early-stage project planning, in-depth project research, market trend analysis, and qualitative assessments. The web-scalable architecture streams domain-curated information, providing a foundation for AI agents to facilitate reasoning and uncertainty elicitation, and supporting future expansions through specialized agents for specific challenges. By systematically integrating AI with industry domain knowledge, this work not only enhances efficiency and decision-making in the construction and infrastructure sectors but also lays a foundation for addressing broader government efficiency. We believe this work will contribute significantly to future AI-driven endeavors in this industry and inform best practices in AI Ops, accelerating the transformation of any analog industry to digital workflows.
\end{abstract}


\section{Introduction}

Data and technology have historically revolutionized industries by enhancing efficiency, productivity, and profitability. Sectors such as advertising, banking, healthcare, and real estate have undergone significant transformations. Yet, the global construction industry—a foundational sector that shapes our physical world—remains a notable exception. Despite being the largest and most capital-intensive industry, directly impacting the planet through extensive infrastructure development \cite{Natureclimate, UNclimate}, it has not fully embraced data-driven transformation. Productivity has stagnated \cite{goolsbee}, and profit margins remain narrow \cite{MCkinsey1, Bahrproductivity}.

At the heart of this inertia lies a critical challenge: the lack of universally available, reliable, and relevant data to inform mission-critical decisions. Conversations with over 100 industry leaders reveal a consensus that this data deficiency is the root cause of systemic inefficiencies. The industry's global sprawl and complexity have allowed it to sidestep traditional regulatory jurisdictions, resulting in fragmented, dispersed, and non-uniform data practices. Unlike other industries, there are no standardized requirements for reporting activities, retaining records, or even common formats for information dissemination. This lack of standardized and accessible data is compounded by the pivotal role of government as a primary driver of infrastructure spending, further complicating transparency and efficiency in decision-making.

Government plays a pivotal role as the primary driver of infrastructure construction spending, involving a wide array of products, materials, and services. Government expenditure constitutes a substantial portion of GDP—ranging from around 20\% in countries like Guyana, Ireland, and Niger to over 50\% in France, Italy, and Belgium \cite{IMF2022}. Public procurement alone accounts for approximately 12\% of global GDP, amounting to \$11 trillion out of a nearly \$90 trillion global GDP in 2018 \cite{OECD2021}. However, the complexity of tracking procurement and spending across multiple levels of government—national, state, city, county, metro, road authority, and energy sectors—compounds the challenge. The lack of transparency and data standards hinders efficiency and accountability in government spending.

This systemic issue has profound implications. For example, a construction firm seeking bridge construction opportunities globally faces the daunting task of accessing procurement announcements from thousands of non-uniform sources across different regions. This fragmentation limits the industry's ability to identify opportunities, assess risks, and make informed decisions at scale.

The consequences are evident in the industry's historical inefficiencies. Major infrastructure projects frequently suffer from cost overruns and delays. Honolulu's rail transit line, initially estimated at \$4 billion, ballooned to \$11.4 billion, with a target completion date of 2031 \cite{honolulu}. Similarly, California's high-speed rail project from Los Angeles to San Francisco, originally budgeted at \$33 billion, is now projected to cost \$100 billion and be completed by 2033 \cite{hoover}. These cases illustrate common issues of cost escalation, engineering challenges, and political obstacles that plague major infrastructure projects worldwide. These inefficiencies are exacerbated by systemic cost escalations, engineering challenges, and delays, reflecting deeper structural issues within the sector.

The stakes are high for all stakeholders. Engineering, Procurement, and Construction (EPC) firms routinely achieve low single-digit margins while assuming undefined risks. Suppliers struggle with demand visibility, leading to supply chain inefficiencies. Government sponsors expend enormous resources during project planning, often resulting in suboptimal outcomes. The lack of a common, accessible archive of industry data limits stakeholders' ability to learn from previous activities, evaluate partners, and properly assess risks. By introducing automation at scale, the system can adapt to regional contexts, ensuring that insights are not only accurate but also globally scalable, overcoming traditional barriers to data standardization in infrastructure projects.

Moreover, the industry faces new challenges that intensify the urgency for transformation. Since 2020, approximately 82.5\% of construction materials have experienced significant cost increases, with an average jump of 19\% \cite{Gordian2023}. The price index for steel mill products more than doubled, soaring nearly 142\% from October 2020 to November 2021 \cite{AGC2021}. These surges, compounded by labor shortages exacerbated by COVID-19-induced retirements and health concerns \cite{Simonson2021}, delay projects and escalate costs, affecting the entire supply chain and economy.

Climate change further complicates the landscape. Extreme heat causes materials like steel and concrete to expand and degrade more rapidly, leading to failures in railways, roads, and power lines \cite{Schlanger2024}. Our infrastructure is simply becoming too hot to function effectively. Adapting to these realities requires transparency and visibility into demand and inventory to navigate future uncertainties. This transparency is critical not only for managing inflation and supply chain disruptions but also for building resilience against climate-related stresses.

Demographic shifts add another layer of complexity. With 41\% of the construction workforce expected to retire by 2030 \cite{PROCORE}, there is a pressing need to counteract impending human capital turnover. Traditionally reliant on human judgment and networks, the industry must now integrate human and AI systems to enhance productivity. Large language models (LLMs) and data-driven AI systems can serve as cognitive capital, preserving and building upon the knowledge of experts even as experienced personnel retire. These trends underline the growing need for systemic change, driven by advanced technologies that can align operational efficiency with the sector's evolving demands.

The urgency for transformation is underscored by the forecasted \$130 trillion investment in infrastructure upgrades over the next five years \cite{Fuchs}. This unprecedented investment creates both opportunities and challenges. Companies and governments must adopt advanced technologies to maximize this investment efficiently. Competitive pressures and historically low margins will drive stakeholders to embrace these technologies or risk being left behind.

Addressing these multifaceted challenges requires a comprehensive approach that leverages advanced technologies, integrates cognitive capital, and establishes clear data standards. We introduce a dual-layered AI approach to tackle these issues: the foundational \emph{Mechanical AI} and the advanced \emph{Thinking AI}. Mechanical AI centralizes and structures raw data into a predictive toolset, bridging knowledge gaps and enhancing clarity across the sector. Thinking AI leverages this structured data to apply expert reasoning to multi-modal data from project reports, public procurement documents, forward-looking plans, and systematic web-scale information to inform strategic decision-making.

At the core of this technological leap is the \emph{Data Mesh}, a vast infrastructure dataset integrating diverse data sources and constantly updating. This mesh standardizes information, enabling a wide range of analyses and insights within construction entities and related fiscal research and government-entity analysis. The Data Mesh tracks over across 1.5 million project records, 27 million tender records, and 600,000 brownfield assets, including alternative data sources like PDFs, planning documents, and scalable web content. It also houses 10 billion records related to risk activities and construction engineering research sources, revealing nearly 2.3 million new entities. These efforts establish robust data standards for an industry historically devoid of such norms, significantly enhancing foundational visibility.

\begin{figure}[ht]
    \centering
    \includegraphics[width=0.4\textwidth]{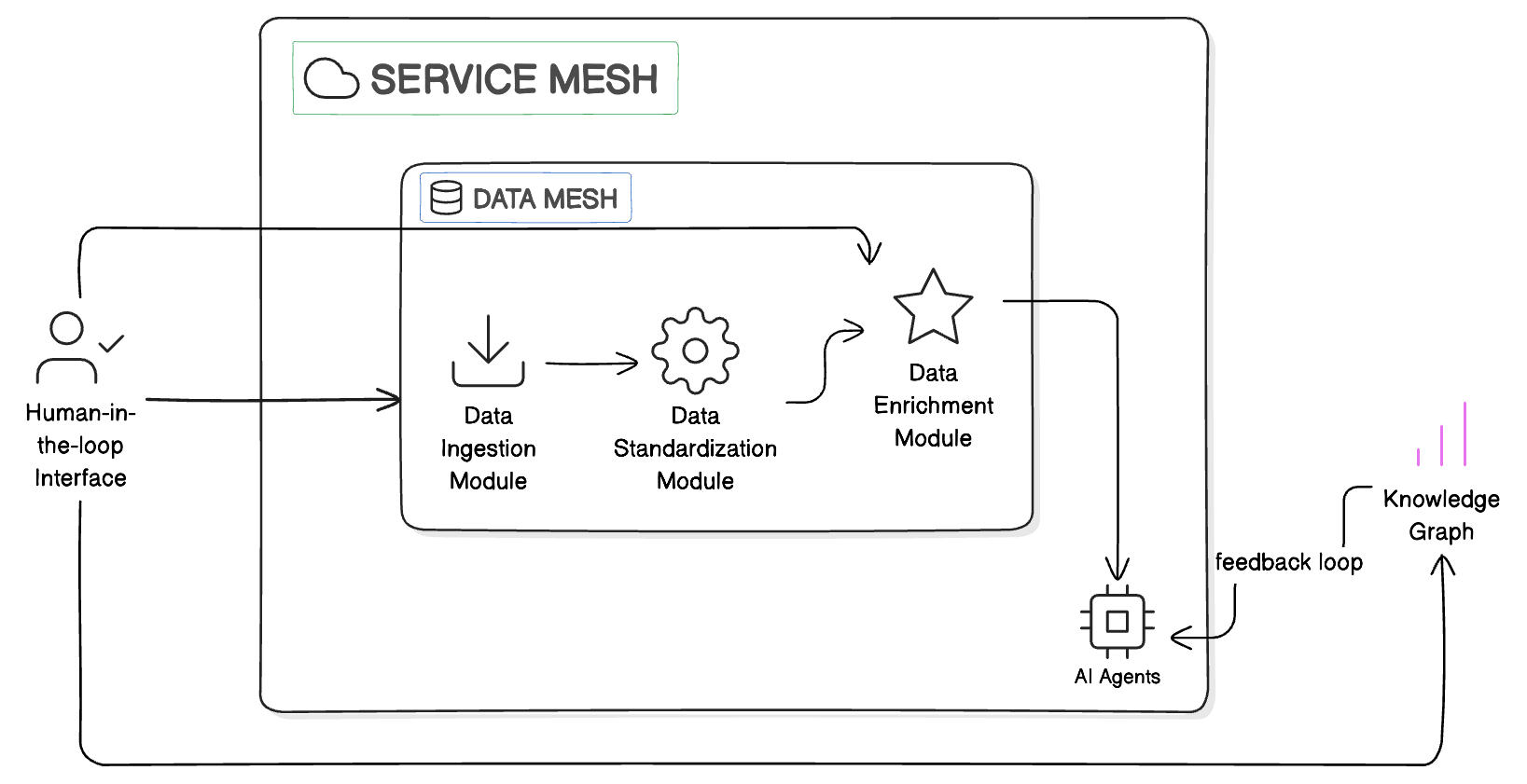}
    \caption{Simplified Abstraction of Software Ecosystem Architecture}
    \label{fig:fig1}
\end{figure}

Figure \ref{fig:fig1} provides a high-level overview of the ecosystem, emphasizing its automation-first approach. Human-in-the-loop processes are employed strategically, focusing on data sourcing, domain knowledge enrichment, and ensuring scalable, high-quality data standards. This balance of automation with selective human oversight ensures robust, domain-informed AI scalability—an ideal mix that leverages human expertise where it adds the most value while maintaining full system automation for efficiency and scalability. Each subsequent section delves into these components in detail, explaining their roles and contributions to the overall system.

By systematically integrating AI with industry domain knowledge, our work enhances efficiency and decision-making in the construction and infrastructure sectors. It also lays a foundation for addressing broader government efficiency. We believe this work will contribute significantly to future AI-driven endeavors in this industry and inform best practices in AI operations, accelerating the transformation of an analog industry into digital workflows. This data-driven approach allows stakeholders to identify opportunities faster, reduce inefficiencies, and gain a competitive edge in an industry where profit margins are historically low. As the construction industry faces new challenges such as climate change and workforce attrition, this framework serves as a future-proof solution, combining real-time data analytics with cognitive AI capabilities to address these pressing concerns while driving industry transformation. 

The rest of the paper is organized as follows. Section \ref{sec:DataMesh} describes the proprietary Data Mesh architecture with its core units as data products (see \ref{sec:Dataproduct}, overall system-wide enrichments to reinforce data standards, human-in-the-loop processes in addition to automation, and \emph{Arc} to further help with knowledge representation for the industry and systematic web scalability. Section \ref{sec:knowledgegraph} presents the knowledge representation and the central and meta AI agents for general domain-informed knowledge representation and multi-agents that can be expanded to inform more specific use cases in the future. Section \ref{conclusion} concludes with implications for future research and development, including areas for improvement.

\section{ Data Mesh and Service Mesh Architecture}

The evolution of data architecture in recent years has prominently featured data lakes, designed to alleviate data management bottlenecks and enhance decision-making capabilities \cite{ladley}. However, the centralized nature of data lakes has exposed significant limitations, especially in handling the proliferation of data sources and the increasing demand for timely analysis and processing \cite{datamesh}. To address these challenges, the Data Mesh paradigm has emerged, advocating for a decentralized approach where data ownership is distributed across domains, coupled with federated governance to oversee data integrity and accessibility \cite{datamesh, li2024empowering}.

Data Mesh posits that decentralizing data ownership to domain-specific teams—those who best understand the context and use cases of their data—enhances data quality and relevance \cite{datamesh}. This paradigm shift is particularly relevant in the context of infrastructure construction, where data is often fragmented and siloed across various entities and jurisdictions \cite{li2024empowering}. The integration of federated learning strategies within Data Mesh architectures facilitates the generation of robust data products without the need for sharing raw data, thereby preserving data privacy and security \cite{kairouz2021advancesopenproblemsfederated}.

The adoption of Data Mesh and AI in infrastructure construction offers broader implications for good practices in data management --- a foundational step to build levels of intelligence for this industry. The decentralized nature of Data Mesh promotes data democratization, enabling domain experts to directly contribute to data quality and relevance \cite{datamesh}. This model can be extended to other industries facing similar challenges with data fragmentation and siloed information \cite{li2024empowering, bode2024avoidingdatamessindustry}.

Moreover, the use of federated learning and AI within Data Mesh architectures represents a convergence of cutting-edge technologies that improve data security, privacy, and utility. By decentralizing data ownership and integrating advanced AI capabilities, organizations can achieve greater transparency, efficiency, and strategic foresight in their operations \cite{kairouz2021advancesopenproblemsfederated, goedegebuure2024datameshsystematicgray}.

\textbf{Complexity of Foundational Data in Infrastructure  Construction}

The advent of the digital era has ushered various industries into the realm of data-driven decision-making. However, unlike e-commerce or social media that thrive on standardized data ecosystems, the construction and infrastructure sectors face unique, complex data challenges.

\textbf{Variable Data Volume and Lack of Standardization}
Infrastructure projects exhibit a wide range in scale and scope, resulting in variable data volumes. For instance, a minor local road repair project contrasts sharply with the data generated from constructing a multi-billion-dollar bridge. The absence of a centralized or standardized platform similar to MLS or Redfin in real estate exacerbates this challenge, underscoring a sector characterized by disparate, fragmented, and decentralized data sources. Standardization of these data, particularly in defining key primary and secondary attributes, emerges as a fundamental hurdle.

\textbf{Diverse Update Frequencies}
The update frequencies of data sources within the infrastructure domain vary significantly. Procurement data might be updated daily, while foundational project details could undergo updates quarterly or annually. This variation complicates the timely analysis and integration of data, hindering the ability to make informed decisions.

\textbf{Inconsistent Data Fields}
Data inconsistencies between different agencies, countries, or private entities manifest themselves in varied fields, formats, and terminologies. Such discrepancies challenge effective data integration and analysis, necessitating a robust approach to standardize and aggregate data for coherent use.

\subsection{Data Mesh}
\label{sec:DataMesh}
The foundation of our solution is the proprietary data-mesh, which represents the re-creation of foundational data for the global industry.  Building the Data-mesh involved three steps: 1) collecting original source data and autonomously standardizing it, 2) enriching the data via classical ML, genAI and LLM in a web-scalable manner, and finally, 3) tagging, standardizing and organizing it for scale.

\begin{figure*}[ht]
    \centering
    \includegraphics[width=\textwidth]{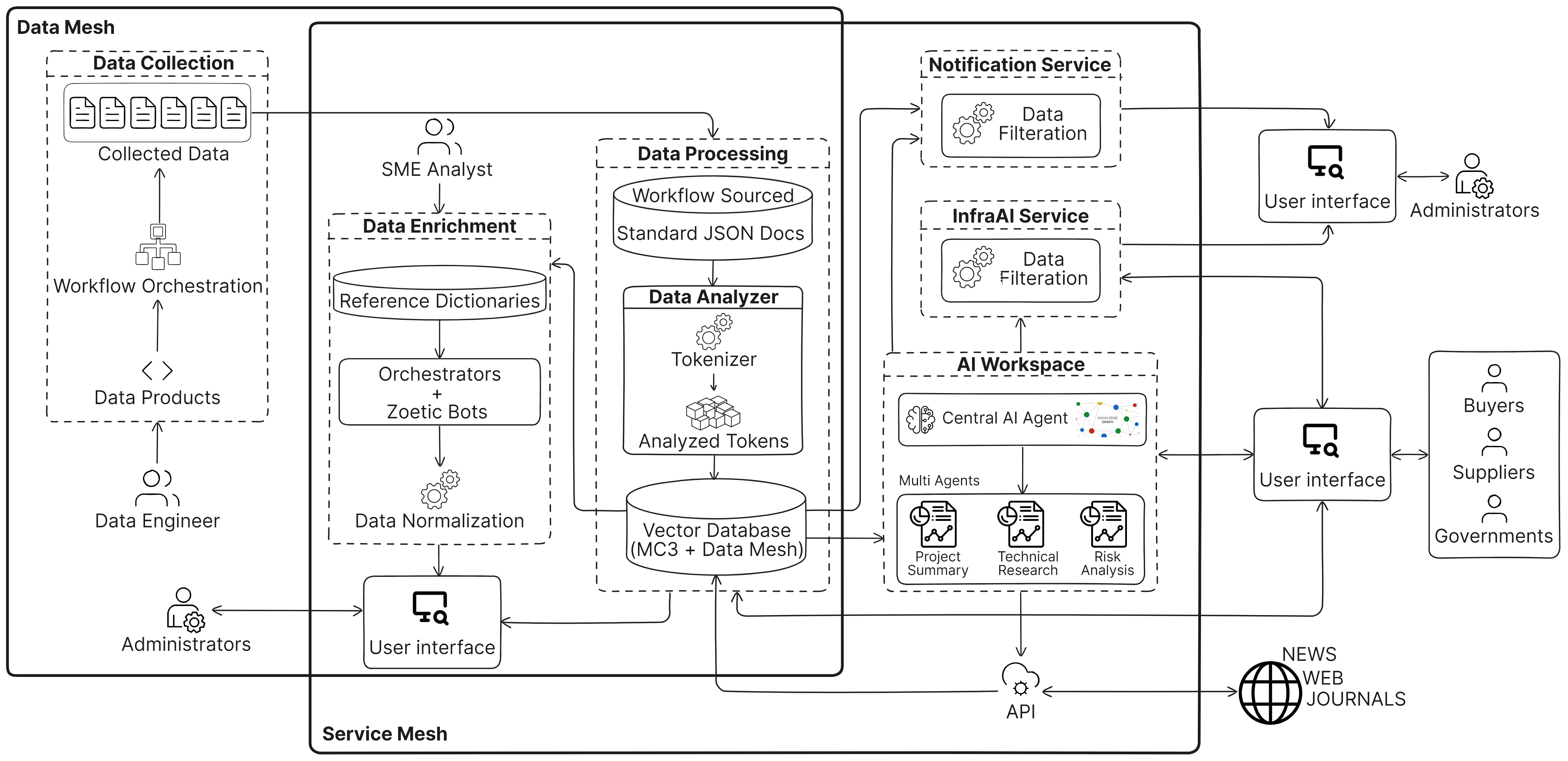}
    \caption{Software Ecosystem Architecture including major System Components}
    \label{fig:systemarchitecture}
\end{figure*}

\textbf{Collecting Original Source Data}
We have created a proprietary process to extract the infra-specific information from over 10,000 distinct global project websites.  Currently, the world’s construction data (opportunities, conditions, participant records, risks, outcomes, etc.) is distributed in every corner of the world through tens of thousands of individual governments. multilateral and private sources. aggregates these records from disparate sources on a daily basis.  The categories of data sources collected and standardized include:

\begin{itemize}
\item Official Government Project Sources
\item Public Procurement Sources
\item Third Party Project Sources
\item Alt Government Data
\item Colleges \& Universities
\item Public Company Filings
\item Engineering Journal Publications
\item Activities, Conditions, and Risks Data Sources
\item Capital Improvement Plan PDFs
\end{itemize}

The types of data collected from these sources typically include project and tender announcements, sponsoring agency, project status, participating stakeholders, budgets, timelines, etc. 

\textbf{Enriching Data via AI}
Much of the industry’s data is not easily found.  The lack of global standards for publishing opportunities or reporting activity means that there is wide variability in the kind of data that is reported.  Since not all official sources include the same information,will augment missing information where possible by employing knowledge graphs and proprietary LLM methods to augment data and improve completeness.

\textbf{Standardizing, Tagging and Organizing}

Once the data are aggregated and enriched, every global record is standardized, tagged, and organized to essentially create one harmonized Data Mesh ready to be mined at scale.

Our Data Mesh represents the largest standardized repository of prospective project and procurement records in the world, all curated to identify business opportunities or inform critical decisions around risk, due diligence or market activity.

At the heart of ’s innovation lies its proprietary Data-Mesh architecture, a decentralized framework designed to revolutionize data collection, processing, and accessibility in construction and infrastructure projects. This architecture amalgamates standardized datasets from a plethora of decentralized sources, providing a unified, accessible endpoint via REST APIs. The design philosophy behind this approach emphasizes flexibility, scalability, and the facilitation of AI-driven insights, thereby addressing critical inefficiencies in the sector.

 Taiyō.AI employs a monolithic repository to manage its Data Mesh architecture effectively. This repository is meticulously organized to facilitate the development, deployment, and maintenance of data products (DP). Each DP resides within its dedicated subdirectory, encapsulating its source code, deployment scripts, and dependencies. The structure emphasizes modularity and reusability, with components such as web scrapers for data extraction, cleaning modules, geocoding services, and standardization scripts. This approach ensures that each data product can be independently curated, standardized, and maintained, aligning with the principles of Data Mesh and Service Mesh architectures.

\textbf{Code Standards and Quality Assurance}

To maintain high code quality, stringent coding practices are adopted. Data formatting follows a consistent standard to ensure uniformity across datasets. Linting checks are performed to identify and rectify code quality issues, adhering to best practices and enhancing maintainability.

\textbf{The Imperative of Data Standardization and Aggregation}

In the face of these complexities, the importance of data standardization in the construction and infrastructure sector is paramount. Only through the establishment of uniform data standards can the industry hope to undertake meaningful comparisons, conduct trend analysis, and perform accurate forecasting. Furthermore, the sector must prioritize the recency of data, enhance entity extraction capabilities, and facilitate the cross-referencing of projects and entities within a graph-structured database to unlock connected insights.

In essence, addressing the data challenge in AI for industry-specific solutions requires a concerted effort to build, standardize, and maintain a comprehensive data ecosystem. This endeavor not only promises to streamline decision-making processes, but also illuminates pathways to innovation and efficiency in a traditionally data-fragmented sector.

The digital age has propelled various industries into an era of data-driven decision-making. However, while sectors like e-commerce or social media benefit from standardized data structures, industries like construction and infrastructure grapple with more complex challenges.

The challenges outlined above necessitate a novel approach, one that combines modern data architectures with domain-specific knowledge. Enter modular approach, rooted in Data Mesh and Service Mesh concepts.

\textbf{Data Mesh \& Service Mesh:} At its core, the Data Mesh philosophy emphasizes decentralizing data ownership and architecture, allowing domain experts to curate and maintain their data products. Service Mesh, on the other hand, focuses on the inter-service communication, ensuring reliable data flow across modules6.

\textbf{Benefits of Decentralization:} By distributing responsibility and ownership,  ensures that data is continuously updated, refined, and maintained by those closest to the source. This approach not only enhances accuracy but also promotes scalability and adaptability to changes7.

\textbf{Data Products:} Central to system is the concept of “Data Products”. These are modular units of data, curated, standardized, and maintained independently. Each Data Product can cater to a specific aspect of infrastructure – be it procurement data, project details, or external risk factors.

\textbf{Web Scrapers:} To populate these Data Products,  employs Python web scrapers. These scrapers, designed with domain knowledge, extract data from diverse sources, ensuring that it adheres to the platform's standardization protocols. The flexibility of Python, combined with its extensive libraries for web scraping like Beautiful Soup and Scrapy, makes it an ideal choice for this task. LLMs are also used in certain instances where applicable, however this falls in less than 30\% of global sources. 

\textbf{Live GenAI Augmentation:}  leverages the power of generative AI (GenAI) to enhance real-time data processing and decision-making capabilities. This system utilizes a combination of web-scalable technologies and advanced AI to provide the latest information on specific subjects, such as entities or individual construction projects, which may be segmented into sub-components. The platform integrates a proprietary Data Mesh along with a large language model (LLM) pipeline, enabling the creation of domain-curated project summaries and the execution of complex analyses. Through this system, multiple specialized AI agents can be invoked to address distinct tasks or challenges, thereby providing tailored solutions based on live data and real-time updates.

\textbf{Ground Truth:} In grounding the digital conversation, theKnowledge Graph utilizes a blend of proprietary and web-sourced data, ensuring a scalable and reliable foundation for project insights. This graph structures data around identified subjects, as determined by end users, and employs sophisticated algorithms to maintain accuracy and relevance. By integrating and grounding diverse data sets in this way,  ensures that users have access to a robust and dependable knowledge base for making informed decisions. This approach not only improves the reliability of data-driven insights but also enhances the overall usability and effectiveness of the AI system in navigating the complex landscape of infrastructure construction.

The core technical architecture is broken down into the following subcomponents: 

\textbf{Development Space} The Development Space in the  architecture serves as the foundational environment where developers and engineers conceptualize, develop, and test innovative AI-driven solutions. This space is meticulously designed to support high levels of creativity and efficiency, providing the necessary tools and frameworks to facilitate rapid prototyping and iterative testing, which are essential for advancing AI capabilities in infrastructure management.

\textbf{Productivity and Collaborations Setup} The Productivity and Collaboration Setup within  emphasizes seamless interaction among team members across various disciplines. By integrating advanced collaboration tools and platforms, this setup ensures that ideas, data insights, and developmental progress are easily shared and accessible, enhancing team synergy and accelerating project timelines in a distributed work environment.

\textbf{Cloud Automation} Cloud Automation at  focuses on streamlining deployment, scaling, and management of infrastructure resources. Utilizing sophisticated automation scripts and tools, this component reduces manual intervention, minimizes errors, and ensures that resources are optimally allocated based on real-time demands, significantly improving operational efficiency and system reliability.

\textbf{Built Artifacts and Software Components} This aspect of the architecture deals with the stable release versions of software components and other digital assets created during the development process. Built artifacts include compiled code, configuration files, and dependencies that are managed through an automated pipeline to ensure consistency and quality across all stages of deployment and production.

\textbf{Automation Orchestration and Cloud Formation} Automation Orchestration and Cloud Formation are pivotal in managing complex workflows and resource provisioning in the cloud. This component utilizes orchestration tools to automate and coordinate multiple processes, ensuring that the infrastructure setup, deployment, and maintenance are executed systematically and align with organizational policies and project requirements.

\subsection{ Service Mesh}
\label{sec:Servismesh}
The  Service Mesh is a critical platform component designed to enhance the accessibility, performance, and security of the Data Mesh architecture within ’s ecosystem. By encapsulating a network of microservices, the Service Mesh facilitates agile and scalable deployment of services, ensuring that client-facing applications and backend services operate with high availability and reliability. The architecture adheres to an API-first design, enabling seamless interaction across all product interfaces, and incorporates transit-level security (TLS) and robust authentication mechanisms at each service endpoint. This approach not only accelerates the development process but also ensures secure, resilient, and controlled scaling on demand, essential for managing complex data workflows in real-time.

\begin{figure*}[ht]
    \centering
    \includegraphics[width=\textwidth]{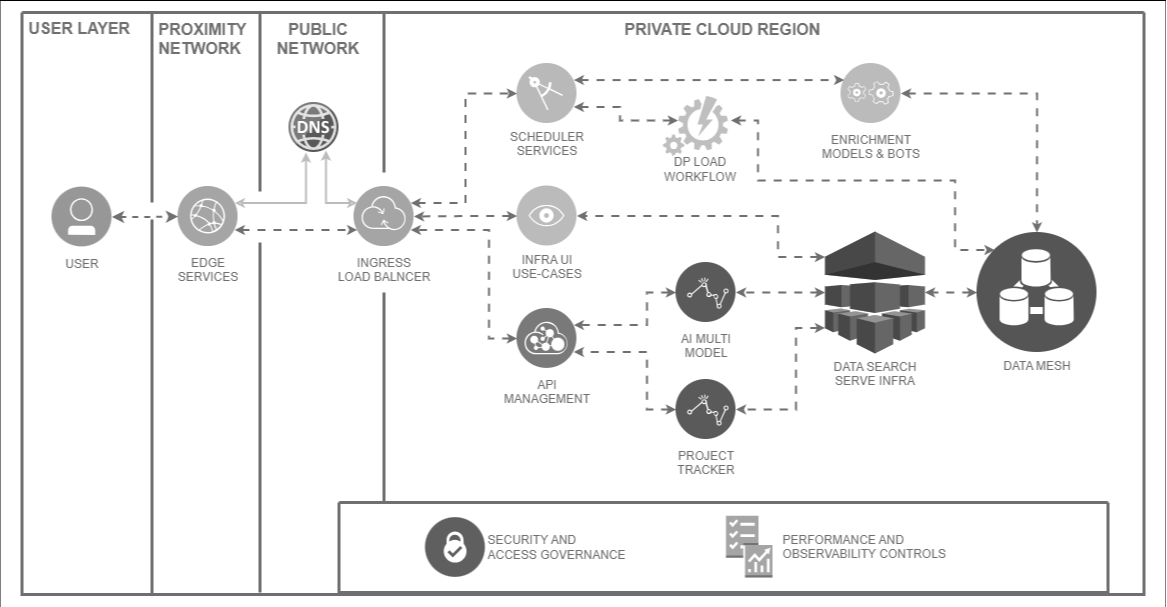}
    \caption{ Service Mesh Architecture}
    \label{fig:service_mesh_architecture}
\end{figure*}

As illustrated in Figure \ref{fig:service_mesh_architecture}, the  Service Mesh integrates several high-level components that work together to deliver comprehensive functionality across the platform. Key components include the Zoetics Model Service, which allows for the live deployment of bots and models without downtime, and the Search/Serve Service, which manages the central data service interface between user interfaces and AI components. These services are designed to deliver both user-filtered and summarized data efficiently, supporting advanced features like geo-hashing and real-time data aggregation.

The DataMesh/Products Backed Service within the Service Mesh enables the integration and management of vast amounts of data, coordinating trillions of JSON documents and data sources to make Data Products available for enrichment, analysis, and AI modeling. Additionally, the Orchestration Service synchronizes various tools and workflows, controlling data product sourcing schedules, and managing internal and external service orchestration tasks.

The implementation of the Service Mesh architecture draws on established service mesh frameworks such as Istio, Kong Mesh, and Linkerd, which offer out-of-the-box service connectivity, zero-trust security, and global observability across all traffic, including cross-cluster deployments \cite{datamesh, zhu2022dissectingservicemeshoverheads}. By integrating these capabilities,  Service Mesh ensures optimal performance and security, with features like automated updates for faster feature delivery, and tools for monitoring application performance at every endpoint. This approach aligns with best practices in modern microservices architecture, addressing the challenges of inter-service communication, traffic management, and system resilience.

The  Service Mesh not only supports the efficient operation of the Data Mesh but also provides a robust, scalable, and secure foundation for the development and deployment of AI-driven applications in the infrastructure construction domain, ensuring that all services operate at peak efficiency and reliability.

\vspace{3.5em}

\textbf{Major Service Mesh Components}

\textbf{Zoetics Model Service}

Zoetics Models are live bots/models along with data products. These zoetics models rules are defined as Enrichment steps and onboarded on platform as live service without downtime and without having to spend extra compute and data movement cost. Data at its local store enriched to desired state and made available for data products search/serve services. 

\textbf{Search/Serve Service}

This is a central service for serving data between UI product interface and AI components. These services not only deliver user filtered data, it also delivers summarized data w.r.t requested dimension and statistical variables. The advanced version of the same services makes geo-hash/geo-grid possible with real time aggregation and filters for users' interest in data.  

\textbf{DataMesh/Products Backed Service}

Master Command Control Center(MC3) delivers configuration and positions of various data products. This service makes possible integrating various json files, folder and meta folders to hold trillions of json/data documents together and deliver source level Data Product and multi source Consolidated Data Products available for Enrichment, Analysis, AI model and deliver via user Use-Cases 

\textbf{Orchestration Service}
	
Orchestration service is a general service for various tools synchronization. This service at present has two versions. A. To control data products souring schedule and trigger various workflows for web sourcing, standardization and ingestions. B. Orchestration of various internal and interface services e.g scheduling bots for enrichment, email and client notifications summaries, and various internal scheduled reports to look at DataMesh and Data Products health summary at point in time.

\textbf{Project Tracker Service}

The Project Tracker Service interacts with the internet and Data Mesh using a knowledge graph to curate infrastructure-related data. It understands domain-specific language to track live statuses from various online sources.
	
\textbf{AI Model Bot Composition Service}

This service connects the knowledge graph with  data, the web, news media, and other sources to construct multi-agent or meta-agent models for advanced AI applications.

\textbf{UI Interface Service for Data Enrichment and Subject Matter Expert (SME)}

An internal tool that supports large-scale enrichment, data standardization, and composition across the entire Data Mesh, facilitating seamless data management.

\textbf{UI Interface Service for Market Landscape}

A user interface service that integrates various components, such as market landscape data, to provide end users with relevant insights.

\textbf{UI Interface Service for Email Draft and Deliver}

This service manages administrative controls for SMEs, ensuring scalable quality validation and enrichment processes with human-in-the-loop checkpoints throughout the system.

\textbf{UI Interface Service for Infra Agent}

A microservice that supports chatbot interactions and delivers on-demand analytical and AI-driven features within the platform.

\subsection{Data Product as a Unit and Their Components}
\label{sec:Dataproduct}
The life-cycle of a data product within the  ecosystem encompasses five critical stages: Scraping, Cleaning, Standardization, GeoCoding, and Ingestion. This life-cycle begins with the meticulous gathering of data across diverse domains—ranging from project specifics and tender details to broader economic indicators and policy landscapes. Each data product undergoes rigorous processing to ensure its reliability and relevance, which serves as a foundational element for AI and machine learning applications aimed at predictive analytics, risk assessment, and strategic planning.

\subsubsection{Lifecycle of a Data Product}

\begin{figure*}
    \centering
    \includegraphics[width=0.9\textwidth]{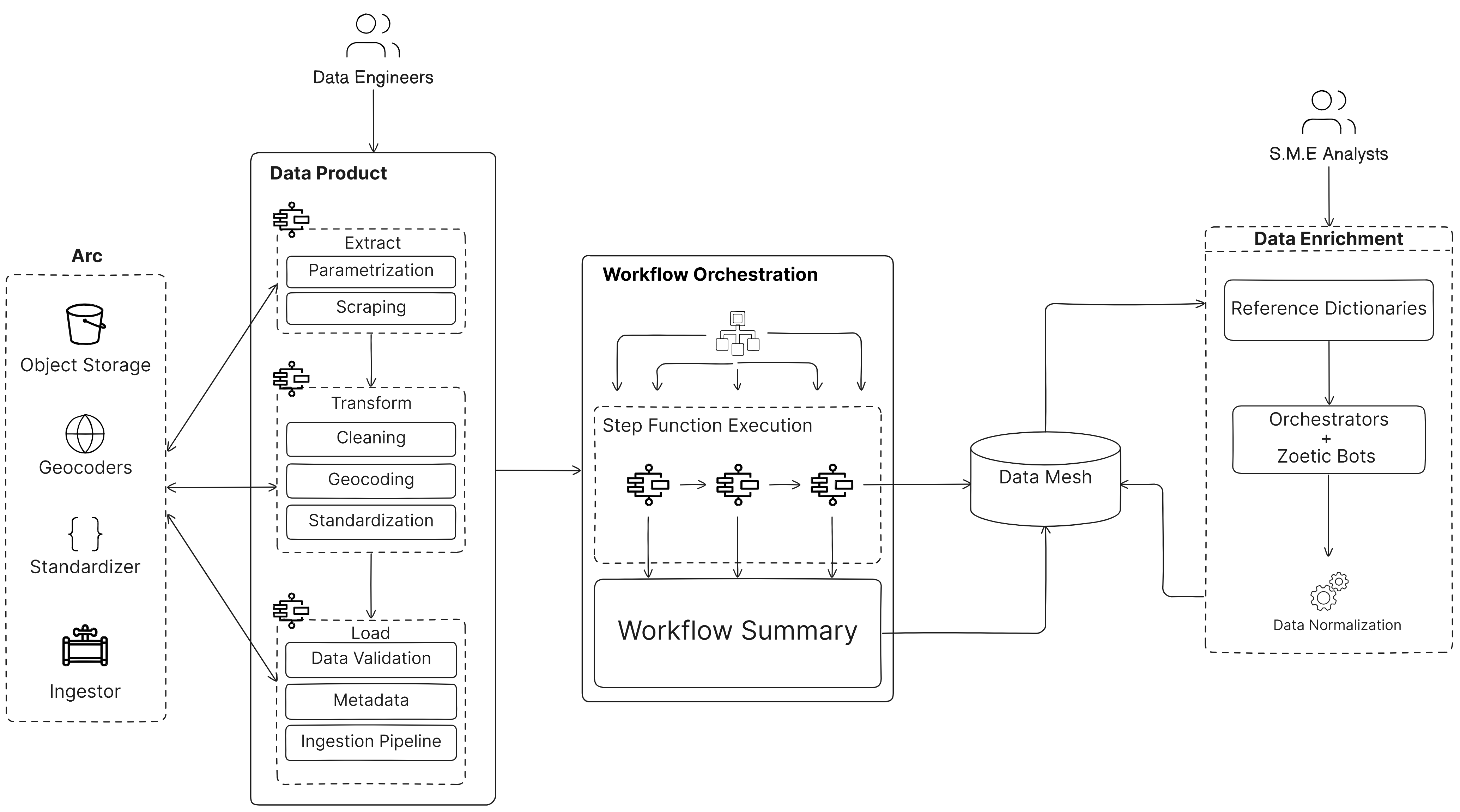}
    \caption{Lifecycle of a Data Product in  Software Ecosystem}
    \label{fig:data_product_lifecycle}
\end{figure*}

The lifecycle of a data product within this software ecosystem is a process involving several stages, from initial data identification to final integration into a Data Mesh. Figure \ref{fig:data_product_lifecycle} illustrates the detailed lifecycle and its various stages. The process begins with the identification of the source of information. Depending on the source, appropriate parameterization and scraping methods are implemented to extract the necessary data. The extracted data is then subjected to a series of analysis functions for cleaning, geocoding, and standardization, which are crucial for ensuring data quality and consistency. 

Once the data product has been prepared, it is deployed to a workflow orchestration tool that schedules and executes the workflow at predefined intervals. During each scheduled run, the data product workflow initiates, starting with the extraction of data using the specified parameterization and scraping techniques to capture only the relevant data. This data is subsequently cleaned, geocoded, and standardized using Arc’s proprietary tool designed for common utilities. 

The next phase involves loading the data into the data mesh, which encompasses three critical steps: data validation to ensure correct data formats and types (e.g., dates in yyyy-mm-dd format), the addition of metadata such as the data product's name and version, and the execution of the ingestion pipeline to incorporate the data into the data mesh. If entries already exist in reference dictionaries, enrichment is applied automatically; otherwise, they appear as DELTA entries, which the enrichment team handles for configuration and updating reference dictionaries. 

The enrichment process is carried out via Zoetic Bots and Orchestrator, framework for batch jobs. The enriched data is then updated in the data mesh. Intermediate data is stored in Object-Storage using Arc in Parquet file format, ensuring efficient storage and retrieval. This systematic engineering approach ensures that data products are consistently high-quality, well-integrated, and ready for use in various analytical and operational applications.

\textbf{Development and Implementation }

\textbf{Technical Foundation and Environment Setup}
To engage with ’s data products, stakeholders are equipped with a comprehensive guide covering the necessary tools and software. The setup process is streamlined through detailed documentation, facilitating easy access to database, application interfaces, and production servers, thereby ensuring a seamless integration into users' local environments.

\textbf{Operational Workflow and Data Engagement}

The operational workflow within the framework is defined by a clear, structured approach to running data projects. Developers and data scientists are go through the process of cloning the project repository, setting up virtual environments, and managing dependencies. This process is complemented by a robust development environment that emphasizes code standards, logging practices, and data format consistency, ensuring that all interactions with the data mesh are both efficient and standardized.

\textbf{Data Validation and Publication}

Following data ingestion, each data product undergoes a stringent validation process, ensuring that attribute mappings and units are accurately reflected before publication. This crucial step guarantees that the data not only maintains high integrity but also aligns with the specific needs and standards of the infrastructure and construction sectors. The validated data products are then seamlessly integrated into dashboard, providing stakeholders with intuitive access to critical insights and analytics.

\textbf{Production Workflow and Deployment}

\textbf{Preparation for Production Deployment:}
The transition from development to production involves ensuring DP-specific configurations are correctly set.

\textbf{Data Products Workflow Orchestration:}

The DAG delineates the cron job workflow, detailing the scheduled execution, resource allocation, and environment variables necessary for the data processing tasks. This configuration facilitates automated, reliable data flow across modules, leveraging Argo Workflows to manage tasks such as metadata scraping, data cleaning, and ingestion into the database.

\textbf{Creating and Managing Merge Requests:}

The final step in the production pipeline involves creating a merge request to integrate changes into the main codebase. Developers push their branch to the repository, followed by creating a merge request. This process underscores commitment to collaborative development and continuous integration/continuous deployment (CI/CD) practices, ensuring that the Data Mesh system remains robust, scalable, and up-to-date.

The development and operationalization of the  Data-Mesh architecture represent a significant leap forward in the application of AI and data science within the construction and infrastructure sectors. By providing an unparalleled level of access to standardized, actionable data,  empowers stakeholders to make informed decisions, optimize project outcomes, and drive a new era of efficiency and success in their projects.

\subsection{Enrichment Workflows, Data Standards and Data Integrity Reinforcement}

\textbf{Enhancing Data Integrity and Relevance}

The process of data enrichment within the  ecosystem represents a innovative advancement in the standardization and augmentation of infrastructure and construction data. This novel procedure leverages offline processes to enrich data at scale, seamlessly integrating with elastic servers. The enrichment process involves augmenting existing datasets with new attributes or refining current data attributes about project or procurement records.

This automation-driven approach ensures that all data within the  ecosystem is enriched daily, significantly enhancing the depth, accuracy, and utility of the information available.

\begin{figure*}[ht]
    \centering
    \includegraphics[width=\textwidth]{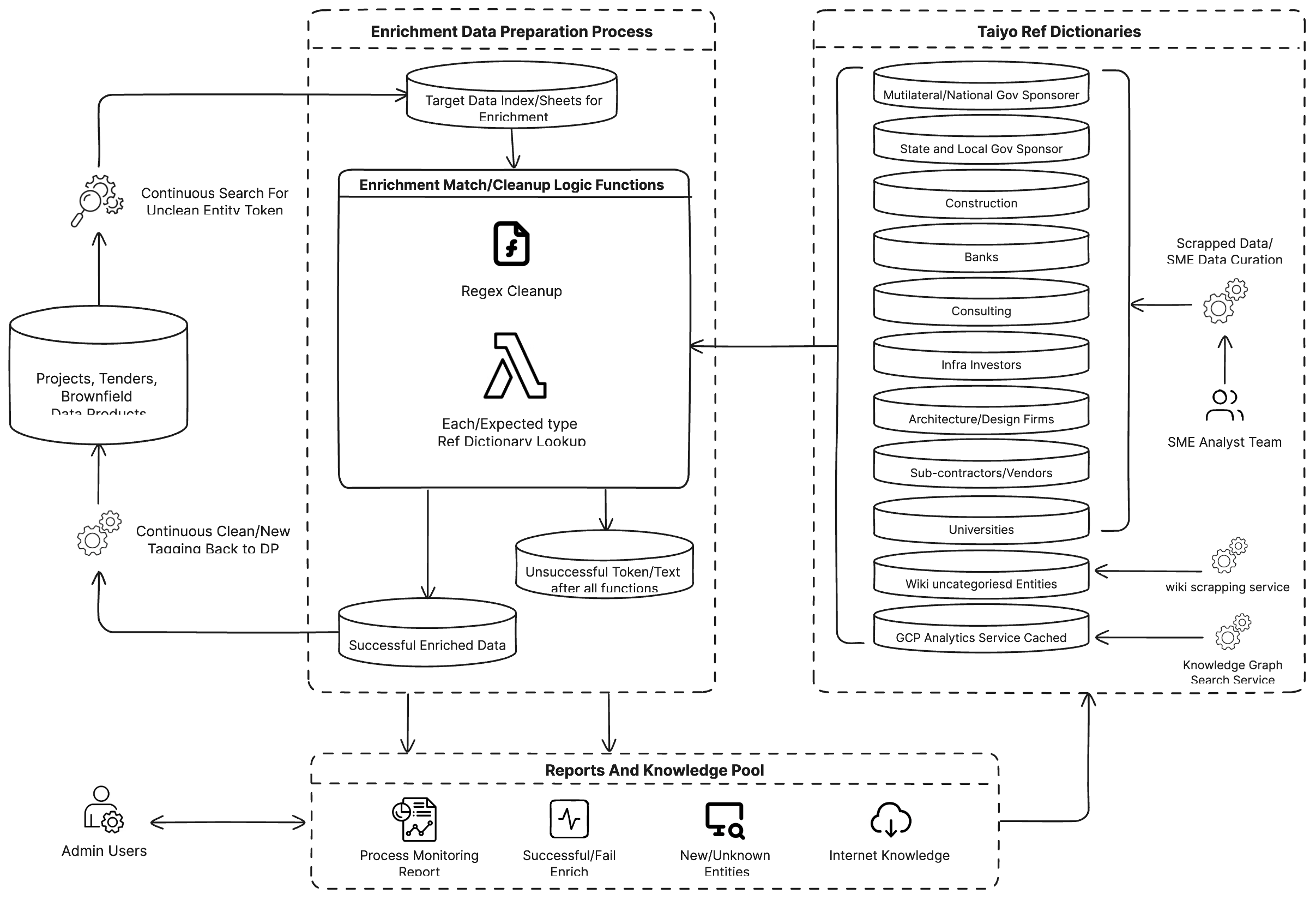}
    \caption{The Enrichment Process to Reinforce Data Standards and Data Augmentation System-Wide}
    \label{fig:enrichment}
\end{figure*}

\textbf{Technical Overview of the Enrichment Process}

The enrichment process is designed to automate the augmentation of data attributes from various internal and external sources. By employing sophisticated algorithms and internal applications,  identifies and incorporates relevant additional information, such as entities \& contact, sector \& tags. This process also involves the critical task of identifying and merging duplicate records, thereby ensuring the uniqueness and reliability of the data. Furthermore, specialized tags and categorizations, including sector \& tags, are applied to facilitate advanced data analytics and insights.

One of the most challenging aspects of this process was the standardization of disparate data sources in a unified manner. The diversity of data formats, terminologies, and standards across sources presented a significant hurdle. 

However, through the development of a robust internal application framework,  has effectively automated the enrichment process. This framework dynamically adapts to varying data structures and standards, enabling the seamless integration and standardization of data.

\textbf{The Significance of Enriched Attributes}

The importance of enriched attributes within the  ecosystem cannot be overstated. These attributes play a pivotal role in enhancing the granularity, relevance, and standardization of data, thereby empowering stakeholders with comprehensive and actionable insights. For instance, the enrichment of country and status attributes ensures that users can perform geo-specific trend analyses and project status assessments. Similarly, the augmentation of budget information and the introduction of sector \& tags enable more precise financial planning and sectoral analysis.

In addition, automation of the enrichment process and its daily execution within the  ecosystem underscores the commitment of the platform to maintain the most current, accurate, and enriched dataset in the global construction and infrastructure sectors. This continuous enrichment process not only addresses the longstanding challenge of disparate data standardization but also opens new avenues for research, analysis, and decision-making in the industry.

\subsection{Human Intervention and Quality Control}

\textbf{Human-Centric Systems:} We focus on developing human-centric systems to ensure control, reliability, and safety of records, enabling robust oversight and operational security.

\textbf{UI End Points for Standardization:} Interactive UI endpoints facilitate the systematic standardization of diverse data attributes, such as project status updates from numerous sources, ensuring consistency and reliability.

\textbf{Outlier Analysis for Quality Assurance:} Through outlier analysis, the system identifies and addresses data anomalies, enhancing the quality and accuracy of the information.

\textbf{Enrichment of Project Attributes:} The interface allows for the enrichment of project attributes, including detailed methods of project delivery, thereby improving data depth and utility.

\textbf{Refining Search and Filter Capabilities:} Advanced search functionalities allow users to refine sector searches based on free-text queries, improving the relevance and precision of search results.

\textbf{Human Curation at Scale:} Extensive application tools and offline administrative capabilities enable human operators to collect data features and maintain high standards on a scale, ensuring the integrity and effectiveness of the system.

\subsection{Advancing Data Structuring with Arc}

In the rapidly evolving landscape of data engineering, managing data pipelines efficiently and ensuring data quality have emerged as paramount challenges.Arc introduces a pioneering approach to tackle these challenges, offering a suite of utility functions, and robust data validation checks. This innovative library streamlines the development and maintenance of data products by integrating key functionalities directly into the data pipeline.

\textbf{Arc}

Arc is architected to enhance the data pipeline's efficiency through its distinct components:

\begin{itemize}

\item \textbf{Utility Functions:} A collection of general-purpose functions designed to facilitate common data operations such as connecting to buffer buckets, reading from, and pushing to Object Storage.
\item \textbf{Validation Check:} Leverages the Pandas library to validate data against predefined standards and expectations, ensuring data quality and integrity.

\end{itemize}

The integration of Utility Functions within Arc offers a seamless approach to data manipulation, significantly reducing the boilerplate code required for data operations. Lastly, the Validation Check ensures that the data products meet the highest standards of quality.

\textbf{Implementing Arc}
A practical implementation of Arc demonstrated significant improvements in the data ingestion process for a large-scale data product. By replacing generic ingestion scripts with Arc's, the data ingestion became more streamlined, with automatic handling of various ingestion parameters and enhanced error handling capabilities.

The integration of Arc into data pipelines involves replacing common utility functions with those offered by Arc, thereby standardizing and simplifying data operations.

 Arc has the potential to influence the way data pipelines are developed and maintained by automating critical aspects of the data management process. Its impact on improving data quality and pipeline efficiency is substantial, offering a promising path towards more reliable and scalable data products. Future enhancements may include broader integration capabilities with other data management systems and further automation of data pipeline components.
 Arc represents a significant leap forward in data pipeline management, providing a robust framework for data ingestion, updating, and validation. By automating essential processes and offering a suite of utility functions,Arc enables developers to focus more on delivering value through their data products and less on the intricacies of data pipeline management. Its adoption not only promises to enhance the efficiency and reliability of data products but also sets a new standard for data pipeline development in the engineering field.

\section{Systematic AI Framework for Real-World Applications}
\label{sec:aiframework}
The systematic AI framework within  is centered around a Central AI Agent complemented by a suite of Multi-Agents. This framework exemplifies advanced data integration and visualization techniques tailored for infrastructure projects. The Central AI Agent utilizes domain-specific language and a modular approach, allowing for a "human-in-the-loop" process that guides the creation of knowledge from standardized AI data and scalable web sources. This knowledge is then linked to various Multi-Agents, each designed for hyper-specific tasks, such as researching the latest project updates or extracting best practices from top-tier journal publications. Other agents, including multi-modal ones, are equipped to understand and analyze local environmental conditions, assess risks, or conduct in-depth comparative research on past projects to inform qualitative judgments.

A key feature of this framework is its focus on uncertainty quantification and transparency. The system continuously calls underlying processes to ensure that results are robust and reliable—this is one of its distinguishing factors. Beyond the deployment of Multi-Agents, the framework includes Meta-Agents capable of real-time market analysis. These Meta-Agents can scan ’s data, news sources, and the web to either compile a comprehensive list of opportunities or provide a completed market landscape view to understand stakeholder dynamics.

The framework's modularity extends to its internal workspace, designed to mitigate risks associated with downtime or latency. This workspace keeps various large language models (LLMs) modular, enabling the system to call upon different models—such as GPT-4o, Mistral, LLaMA, or other configurations like RAG or graphRAG - depending on the specific use case for Multi-Agents.

To illustrate the practical application of this framework, consider the challenge of managing and interpreting the multitude of data associated with large infrastructure projects. These projects often involve wide-reaching impacts, making it essential to accurately integrate diverse data sources such as recent news, internet publications, and bibliometrics. The integration must clarify the project's status, assess risks, identify stakeholders, and visualize its development over time.

The  system addresses this challenge through a novel approach to data aggregation and visualization. By vetting and integrating both official and non-official information from scalable web sources,  provides a dynamic, real-time visualization of a project's timeline and key attributes. This system stands out not only in its ability to manage diverse and complex information but also in its method of standardizing and presenting this information in a user-friendly format. Leveraging advanced algorithms for outlier analysis, the system enriches data attributes such as project delivery methods and stakeholder details from a vast array of sources.

\subsection{Knowledge Graph and Representation}

The  Knowledge Graph embodies a sophisticated structure designed to enhance and structure real-world data for more insightful decision making. Using a triplet architecture that comprises subject, lexicon, and object, this system forms the foundational framework for a highly adaptable and dynamic knowledge graph. Grounded in a diverse array of data sources including the internet, news media, and technical engineering research publications, the graph integrates these elements in a modular fashion, providing a robust platform for augmenting real-world knowledge structuring.

Knowledge Graphs (KG) represent a leap forward in structuring and utilizing data, offering a dynamic and interconnected framework that enhances data accessibility and interoperability. In enterprise settings, KGs facilitate a deeper understanding of complex relationships between diverse data points, enabling more informed decision-making processes \cite{hofer2023construction, 10.1145/3209978.3210085, Guu_2015}. Their application in construction and infrastructure can revolutionize how projects are conceptualized, planned, and executed, by providing a holistic view of project data, stakeholders, and external factors.

\textbf{Implementation Details}

\textbf{Symbolic Subjects:} The graph often begins with a symbolic subject when precise starting points are not defined. For example, selecting a geographical location like California and a sector such as airports may not correspond to a pre-existing RDF node or structure. This symbolic starting point allows for the construction of a predicate-based filter to navigate through related tenders and data, fostering more open-ended explorations within a controlled attribute subset.

\textbf{Literals and Lazy Loading:} To support this dynamic structure, literals are employed within an abstract syntax tree. Much like the triplet itself, the specialty of these literals lies in their 'lazy load' capability; they are not pre-populated across every node or edge but are instead created on demand based on user interactions. This method emphasizes efficiency and relevance, tailoring the data presented to the needs of the user journey.

\textbf{Complex Predicate Equations:} The relationship between the subject and object in the  Knowledge Graph is not merely a straightforward predicate but involves complex equations with variable weightings. This nuanced approach allows for a deeper and more precise interpretation of the data relationships, enhancing the graph’s ability to reflect real-world complexities.

\textbf{System Implementation of Subject/Object:} The implementation of the subject/object system within the  Knowledge Graph involves specific types of literals. These literals vary in definition and application, providing a flexible yet structured way to encode and retrieve knowledge.

\textbf{Lexicon:} Lexical Relation between Subject and Object is consideration of all Subject/Object Properties, their match level/types/algorithms and weighted priorities.

\begin{figure}[ht]
    \centering
    \includegraphics[width=0.5\textwidth]{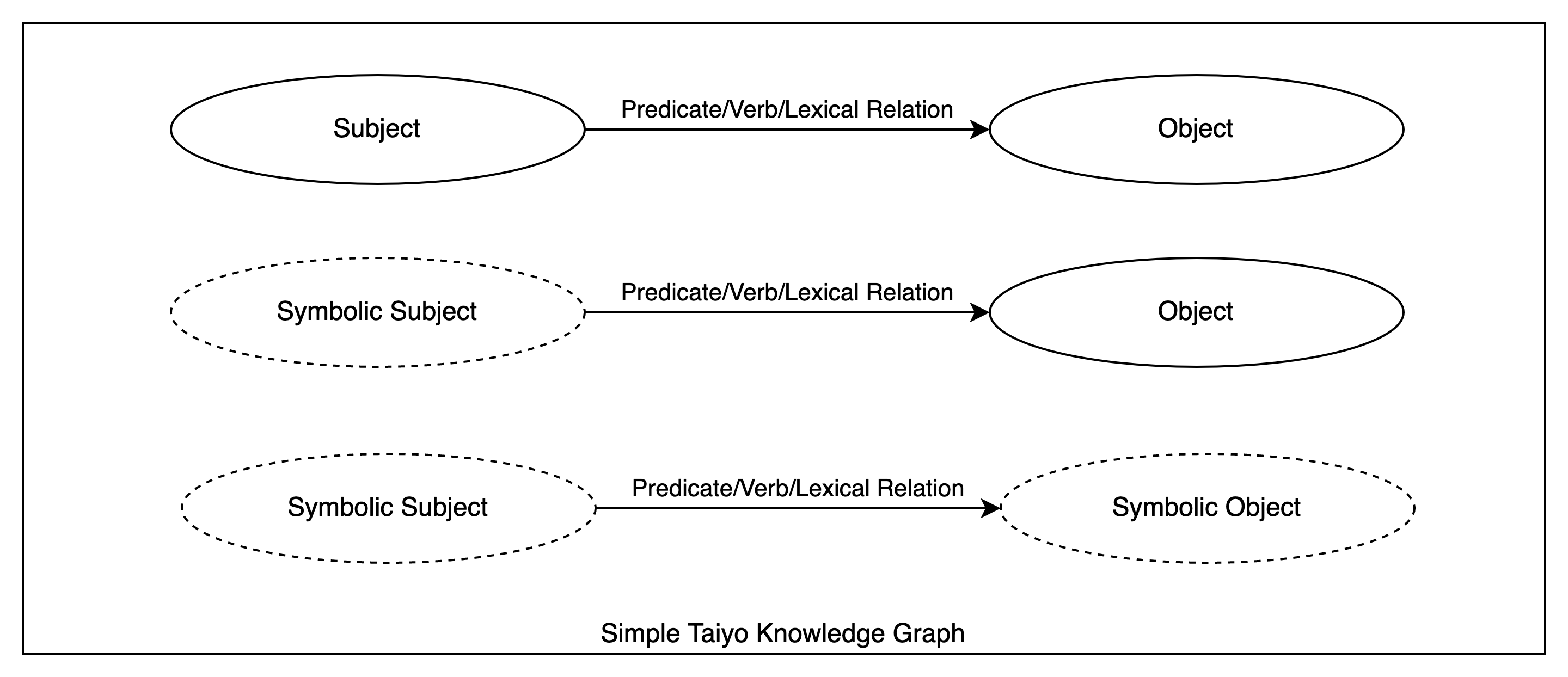}
    \caption{Simple Knowledge Graph Construct}
    \label{fig:KG1}
\end{figure}

\subsection{Knowledge Representation and Handling}
\label{sec:knowledgegraph}

\textbf{Literals}
Literals are fundamental data structures implemented using AST Vectors from Platform. Platform Vectors and its formation in the next chapter. 

\textbf{AST - Abstract Syntax Tree} 
This is technical implementations of various types/data structure and specific use-case implementation for KG Triplet. The  platform implementation of knowledge graph and its specific version is fundamentally based on lazy load and exploration while traversing principles to maintain system performance and user experience. So AST here is specially a Lazy AST based on data vectors. 

\textbf{Subject/Objects}
 Subject/Objects have properties and property values. There are multiple sub definition types for Subject. Relation is defined using Is-A and Has-A notations (explained in next chapter). Collection of Subject/Objects indicated using simple knowledge graph oval shape nodes.  

\textbf{Abstract Subject Types}

Data Point - This is the most generic type and fundamental type and very close to data sourced from the internet. In  Knowledge Graph this is symbolic notation for every start traversal node. For Abstract Subject Type source properties varis from its implementations specifics and title, description, url, location, and timestamp are most commonly used hence available as Abstract Subjects Type instances.

\begin{figure}[ht]
    \centering
    \includegraphics[width=0.4\textwidth]{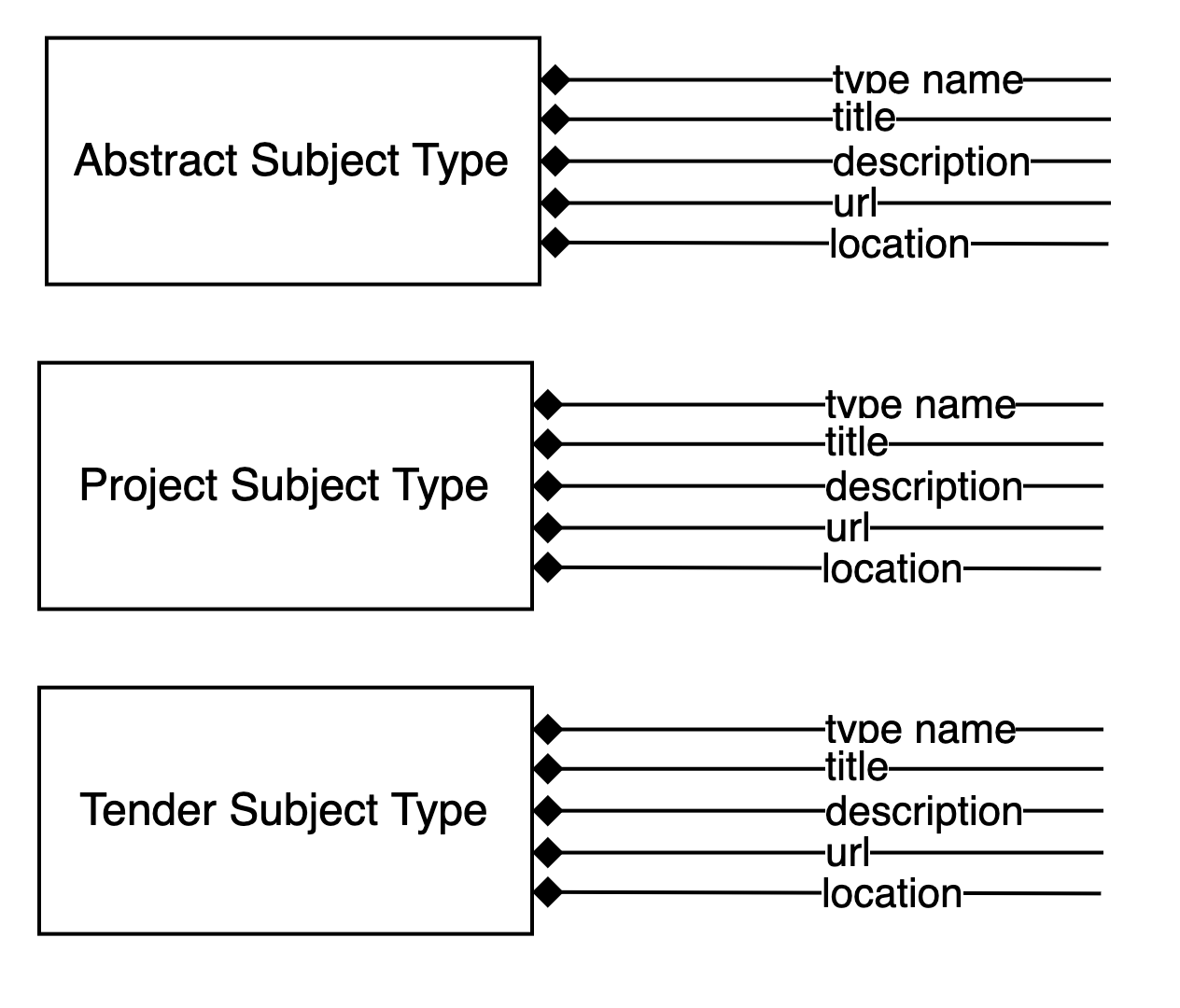}
    \caption{Subject Types}
    \label{fig:Subjects}
\end{figure}

\textbf{Specific Abstract Subject/Object Types}
 
Some specific examples for Abstract Subjects types below. More properties with these types makes a more clearer definition about AST. All traversed/objects associated with the subject via verb/predicate are collections of specific AST. 

\textbf{Symbolic Subject/Object Types}

\begin{figure}[ht]
    \centering
    \includegraphics[width=0.4\textwidth]{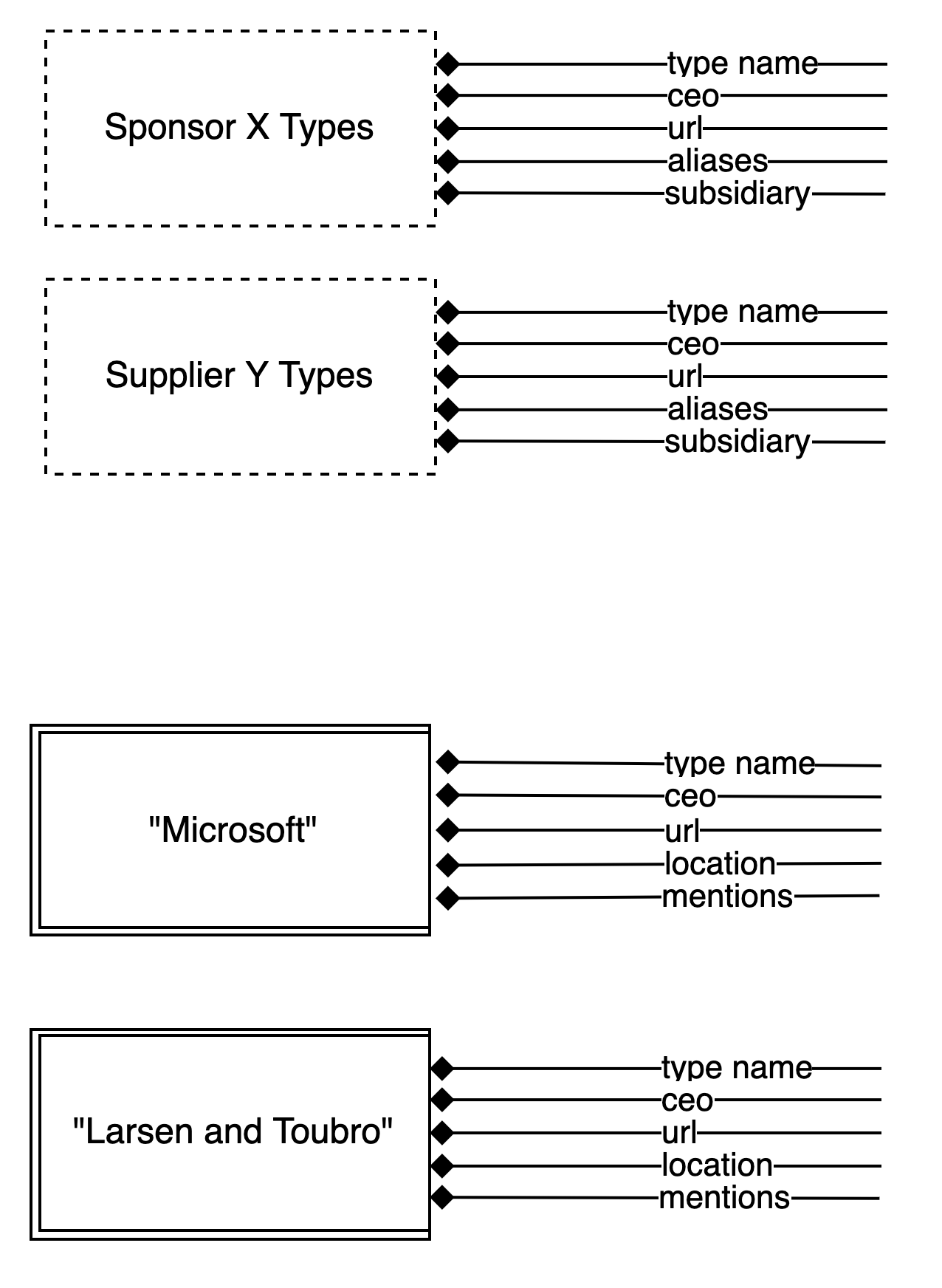}
    \caption{Symbolic Subject Types}
    \label{fig:SymoblicSubjects}
\end{figure}

\textbf{Specific/Term Subject/Object Types}

\textbf{Is-A \& Has-A Notations}
Is-A and Has-A notation helps clear indication between Properties and Literals and Verb.

\textbf{Is-A}
Is-A indicate another literal/nodes are sub type of some generic AST type or AST type collection

\begin{figure}[ht]
    \centering
    \includegraphics[width=0.5\textwidth]{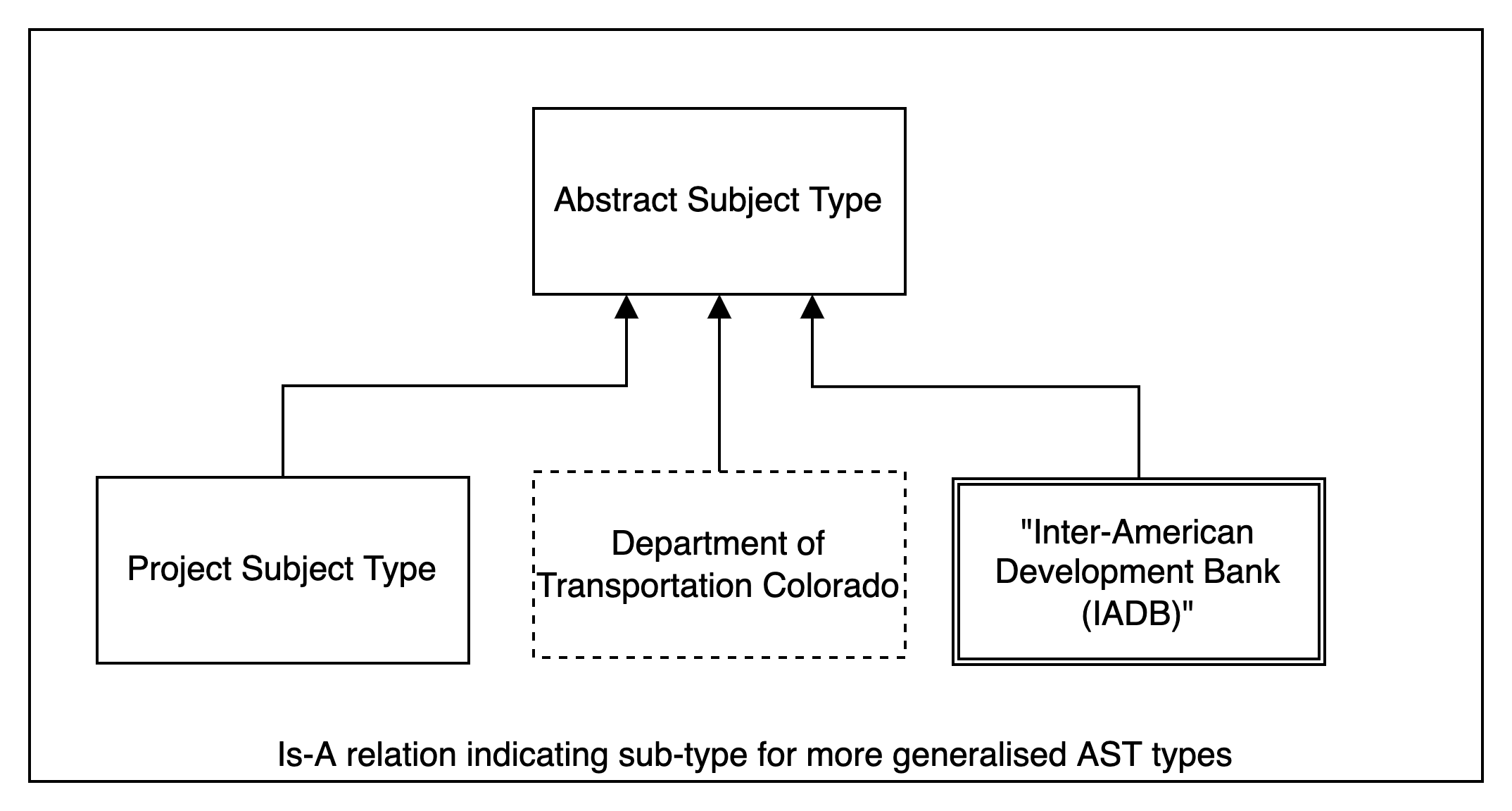}
    \caption{Is-A Relation indicating sub-type for more generalised AST types}
    \label{fig:IsAtype}
\end{figure}

\textbf{Has-A}
Has-A indicate another literal/nodes had independent existence and they are reloaded to other AST type nodes/literals with direct verb or lexical relation.

\begin{figure}[ht]
    \centering
    \includegraphics[width=0.5\textwidth]{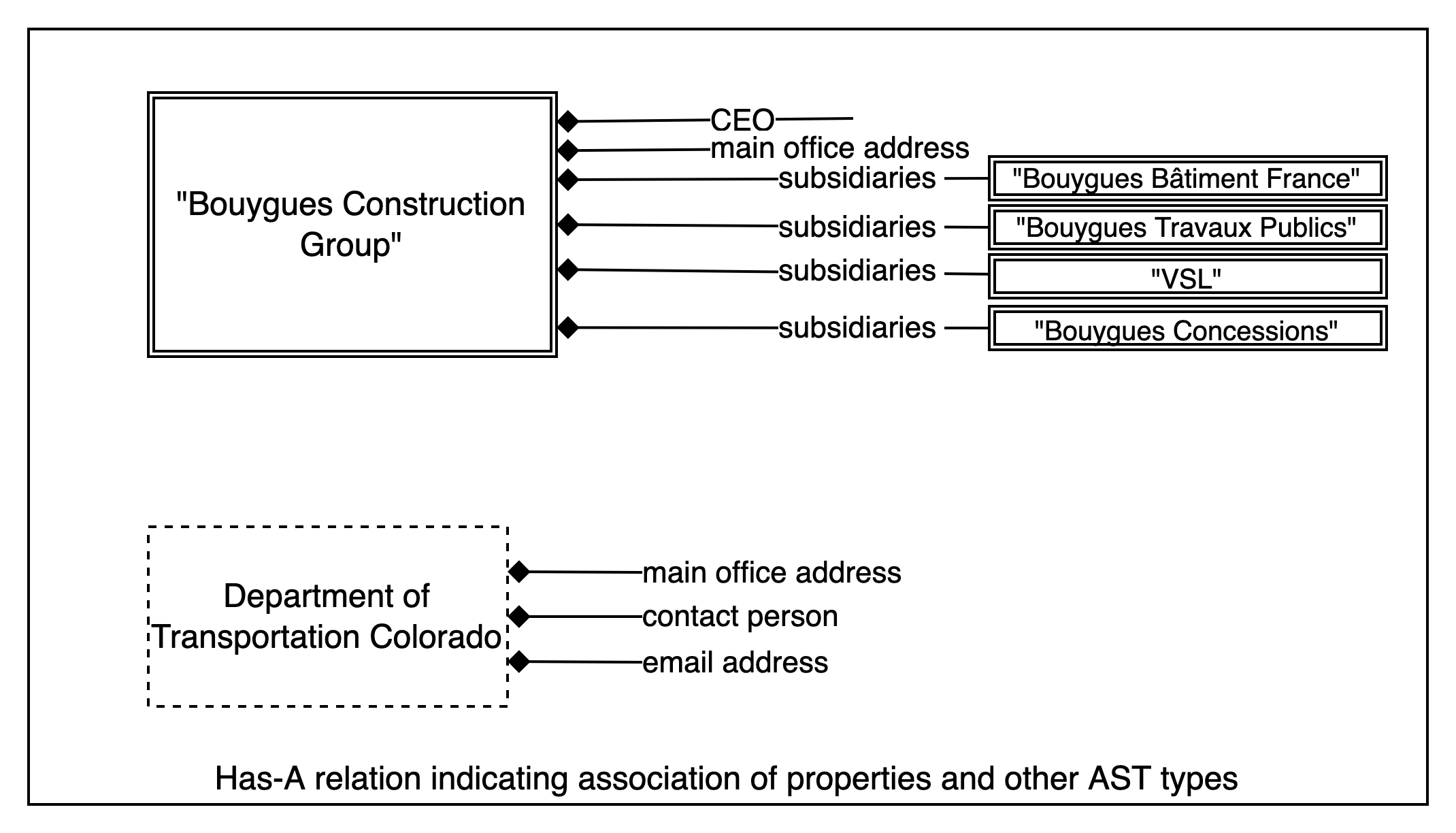}
    \caption{Has-A Relation association properties and other AST types}
    \label{fig:HasAtype}
\end{figure}

\subsubsection{Vector Storage and Graph Traversal}

The vectors are stored/created from Internet sourced data and structured into JSON. Further same JSON structure standardized w.r.t. Infra SME knowledge base information rules. This JSON structure is further analyzed by the English language tokenizer and analyser to store JSON into vector format and yields tokens. Term frequency count on tokens generated by the previous step creates vectors for data retrieval and Lexical Traversal search. 

\vspace{4pt}

\begin{figure}[ht]
    \centering
    \includegraphics[width=0.5\textwidth]{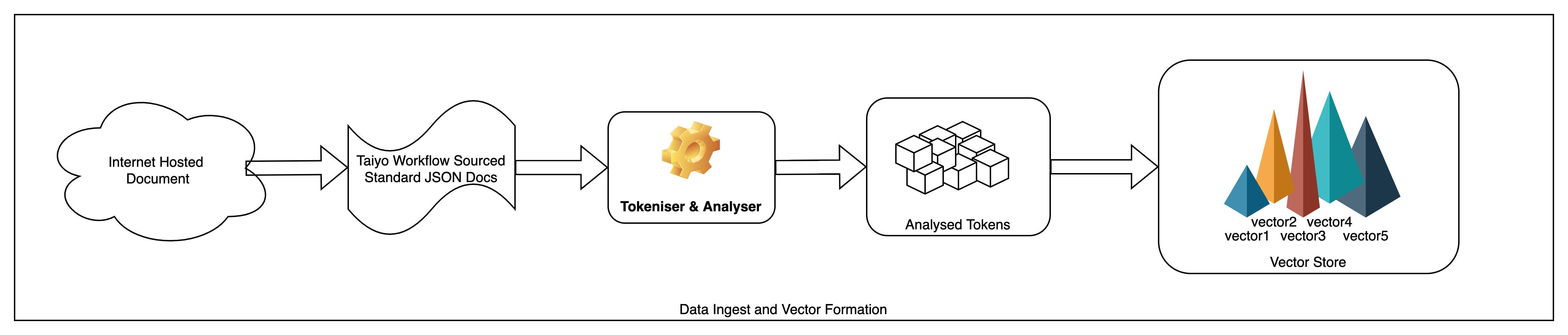}
    \caption{Data Ingest and Vector Formation}
    \label{fig:Vectorform}
\end{figure}

\textbf{Lazy/On-Demand Knowledge Graph traversal and APIs traversal mechanism}
 
 We define Knowledge Graph has an Infrastructure industry specific Subject, Object and Literals and Lexical Relationship defined. This helps various end user use-cases dashboards present and to traverse various nodes in the knowledge base. The APIs interface also gives system administrators controls to define various AST types, infra lexical relations, weighted algorithm definition and interval KG view controls.  

\begin{figure}[ht]
    \centering
    \includegraphics[width=0.5\textwidth]{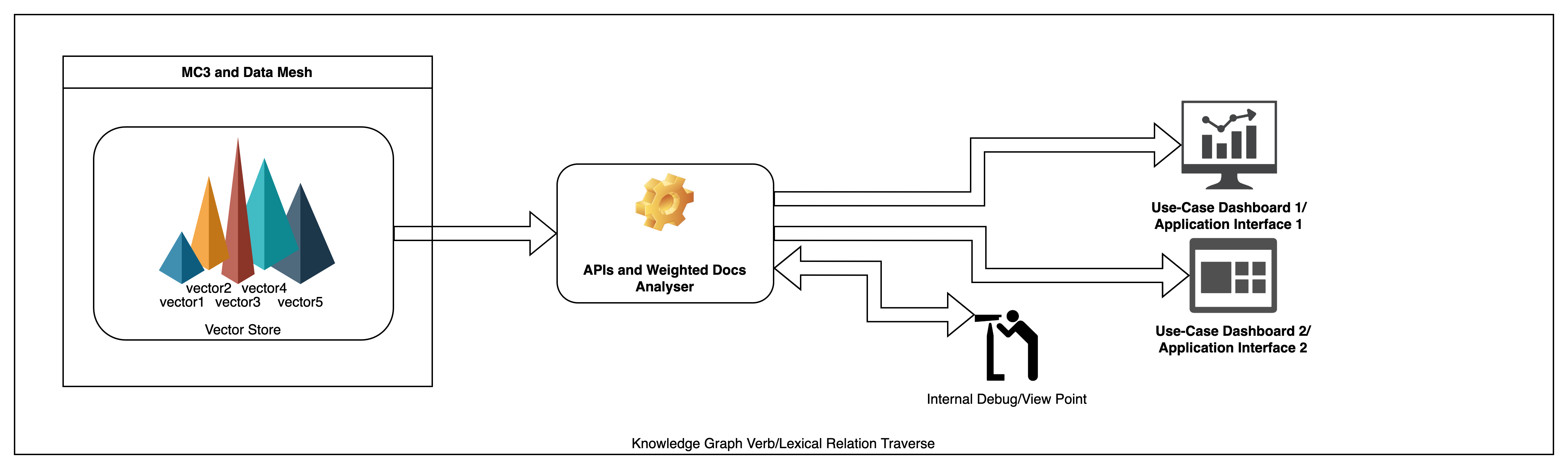}
    \caption{Knowledge Graph Verb/Lexical Relation Traverse}
    \label{fig:KGRelationTraverse}
\end{figure}

\textbf{Verb and Relation Types}
Graph Traversal Predicate --- Verb  in  knowledge base is the next level of relationship definition. Infra Knowledge Graph here is not a very fixed or rigid rules definition. It is more like the English language like syntactic sugar. Here APIs traversePredicate/Verb makes use of term tokens with must, must-not, should limit, more-text-likely free text and many more match definitions and weighted algorithms to choose out of traversed Leterals.

\subsection{Multi-Agent Architecture and Workflow}

\begin{figure*}[ht]
    \centering
    \includegraphics[width=0.9\textwidth]{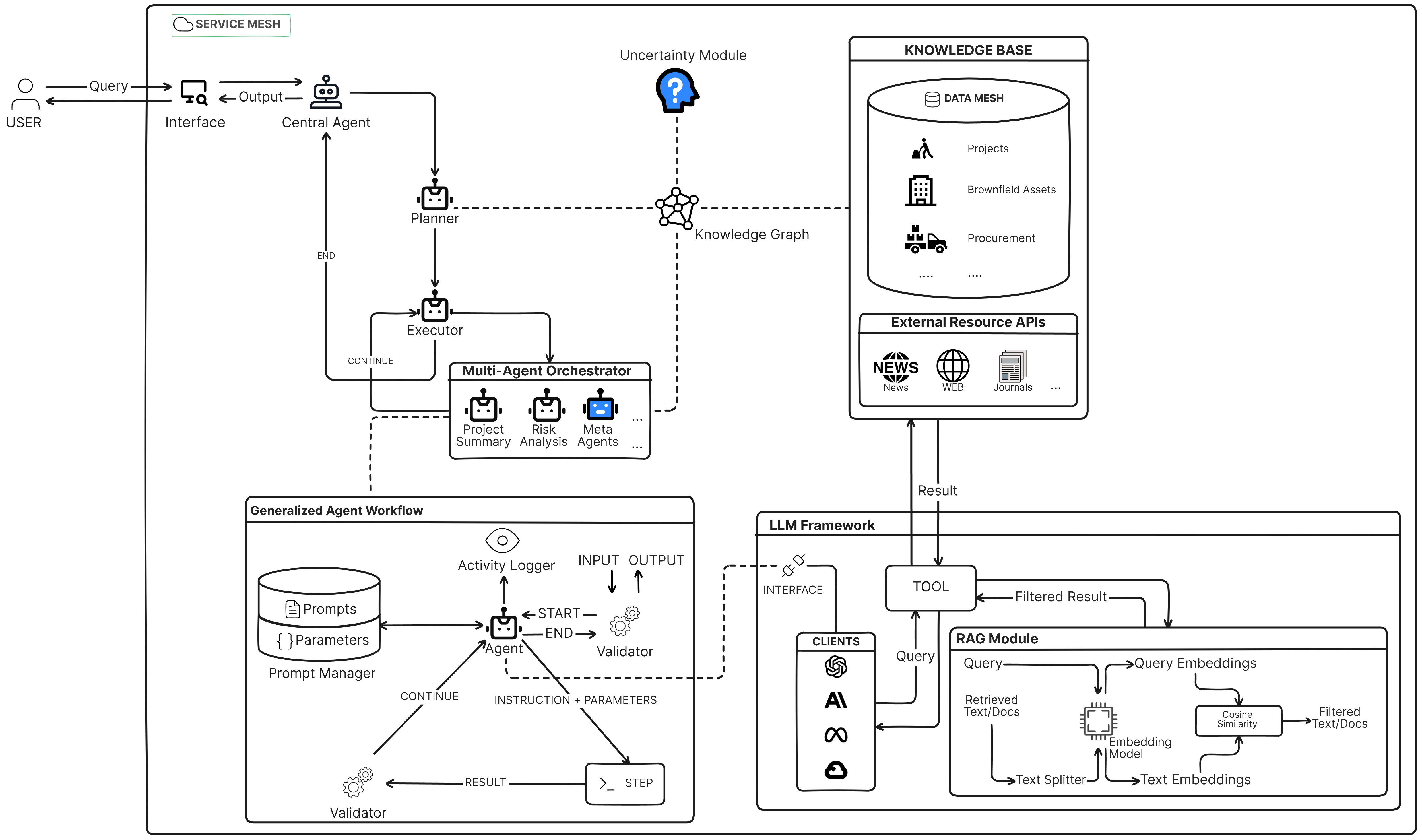}
    \caption{Multi Agent Generalized Workflow}
    \label{fig:multiagent}
\end{figure*}

The proposed architecture introduces a scalable multi-agent framework designed to harness the power of ensemble learning - using machine learning, generative AI, a knowledge graph, and a data mesh for real-world applications. The system is tailored to address the challenges of dynamic, multi-faceted datasets, ensuring relevance, accuracy, and scalability. By incorporating a centralized planning and execution mechanism, uncertainty modeling, and external data integration, the architecture sets a novel standard for AI applications in complex domains such as infrastructure management and construction.

\subsubsection{Modular Design and Workflow Orchestration}

At the core of the system is the \textbf{Central Agent}, responsible for managing conversation context and maintaining the state of interactions. This agent prevents context overload by retaining only essential user queries and derived answers. It interfaces directly with the \textbf{Planner}, which utilizes advanced reasoning techniques such as Chain of Thought (CoT) and Tree of Thought (ToT) \cite{yao2023treethoughtsdeliberateproblem} to develop strategic plans for answering user queries. The Planner has knowledge of available data, resources, and their interrelationships, allowing it to generate multiple potential plans and select the most effective one.

Once a plan is formulated, the \textbf{Executor} acts as an orchestrator, executing the plan step-by-step. It invokes specialized \textbf{Agents} for specific tasks, such as data retrieval, information extraction, or analysis. These agents are designed with components like a Prompt Manager, Validator, and Activity Logger to ensure robust and auditable operations. Each agent is connected to the \textbf{LLM Framework}, which provides a modular interface to various Large Language Models (LLMs) from different providers.

\subsection{Knowledge Graph Integration and Uncertainty Handling}

A key innovation in this architecture is the integration of a \textbf{Knowledge Graph} augmented with an \textbf{Uncertainty Module}. The Knowledge Graph serves as a structured repository of entities, relationships, and attributes pertinent to the domain. It enables efficient querying and reasoning over complex data interdependencies.

The Uncertainty Module addresses several key aspects:

\begin{enumerate} \item \textbf{Data Incompleteness}: Identifies regions or sectors with missing or sparse data, allowing the system to flag areas where additional information is required. \item \textbf{Epistemic Gaps}: Estimates unknowns in the broader context of web and journal knowledge, helping to quantify the limitations of current knowledge and model predictions. \item \textbf{Dynamic Updates}: Incorporates the latest findings from journals, research, and news sources, ensuring that the Knowledge Graph remains current. \end{enumerate}

Mathematically, uncertainty can be modeled using probabilistic graphical models or Bayesian networks within the Knowledge Graph, allowing for reasoning under uncertainty and making probabilistic inferences \cite{pearl, ni2024trustworthyknowledgegraphreasoning}.

\subsection{Advanced Algorithmic Process and Workflow Orchestration}

The \textbf{RAG Module} ensures that only the most relevant information is considered in decision-making processes. It combines embedding-based retrieval with cosine similarity scoring to filter out noisy data and deliver semantically coherent results.

The text processing pipeline includes:

\begin{itemize} \item \textbf{Text Segmentation}: Dividing large texts into semantically meaningful units using algorithms like TextTiling or topic segmentation, which is crucial for handling long documents. \item \textbf{Tokenization}: Utilizing efficient tokenization methods such as Byte Pair Encoding (BPE) or WordPiece to prepare text for embedding. \item \textbf{Embedding Generation}: Employing state-of-the-art models like Transformer-based encoders to generate dense vector representations capturing semantic nuances. \item \textbf{Semantic Similarity Analysis}: Calculating cosine similarity between embeddings to assess the relevance of documents to the user's query. \end{itemize}

Formally, given a query vector $\mathbf{q}$ and a document vector $\mathbf{d}$, the cosine similarity $\cos(\theta)$ is computed as:

\begin{equation}
\cos(\theta) = \frac{\mathbf{q} \cdot \mathbf{d}}{\|\mathbf{q}\| \|\mathbf{d}\|}
\end{equation}

Documents with similarity scores above a certain threshold $\tau$ are considered relevant:

\begin{equation}
\cos(\theta) \geq \tau
\end{equation}

The multi-agent workflow algorithm with uncertainty handling is presented in Algorithm~\ref{alg:multiagent}. This algorithm outlines the step-by-step process by which the system processes user queries, retrieves and filters relevant data, and generates responses while accounting for data uncertainty.

\begin{algorithm}[ht]
\caption{Multi-Agent Workflow Algorithm with Uncertainty Handling}
\label{alg:multiagent}
\begin{algorithmic}[1]
\Require User Query $Q$, Knowledge Graph $KG$, Data Mesh $DM$, Retrieval-Augmented Generation Module $RAG$, Uncertainty Module $UM$
\Ensure Relevant Response $R$
\State \textbf{Initialize:} Central Agent, Planner, Executor, Agents
\State \textbf{Input:} User Query $Q$
\State Central Agent updates conversation context with $Q$
\State Planner generates plan $P$ using Chain of Thought (CoT)
\For{each step $s$ in plan $P$}{}
   \begin{enumerate}
   \State Executor invokes Agent $A_s$ for step $s$
   \State Agent $A_s$ performs the following tasks:
   \begin{enumerate}[i.]
      \item Retrieve relevant data $D_s$ using $RAG.retrieve(Q, KG, DM)$
      \item Assess data uncertainty $U_s$ using $UM.evaluate(D_s)$
      \item Filter data $D'_s$ where $\cos(\theta) \geq \tau$ and $U_s \leq \delta$
      \item Generate output $O_s$ using $LLM.process(\text{Prompt}, D'_s)$
      \item Validate $O_s$ with Validator, considering uncertainty $U_s$
      \item Log activity with Activity Logger
      \item Executor updates context with $O_s$
   \end{enumerate}
   \end{enumerate}
\State \textbf{Return} consolidated Response $R$ from all $O_s$  
\end{algorithmic}
\end{algorithm}

\newpage

As shown in Algorithm~\ref{alg:multiagent}, the system begins by initializing all necessary components and updating the conversation context with the user's query. The Planner component generates a plan using advanced reasoning techniques. For each step in the plan, the Executor invokes the appropriate agent, which performs tasks including data retrieval, uncertainty assessment, filtering, output generation, validation, and logging. The incorporation of the Uncertainty Module ensures that the system considers both the relevance and reliability of the data before generating responses.
The architecture supports a range of agents tailored for specific tasks:

\begin{itemize} \item \textbf{Project Summary Agent}: Tracks developments and extracts summaries for project-specific insights by analyzing timelines, milestones, and progress reports. \item \textbf{Technical Research Agent}: Identifies and synthesizes information from academic journals and technical reports, summarizing best practices and lessons learned. \item \textbf{Risk Analysis Agent}: Evaluates environmental, economic, and macroeconomic metrics using data analytics and predictive modeling to assess project risks. \item \textbf{Comparable Project Agent}: Identifies similar projects using clustering algorithms and similarity measures to learn from past experiences. \item \textbf{Multi-Modal Analysis Agent}: Analyzes time-series data, local area metrics, and macroeconomic indicators to provide comprehensive insights. \end{itemize}

\textbf{Meta-Agents} operate at a higher abstraction level, synthesizing insights across projects, markets, or regions. These agents enable stakeholders to transition from static reporting to dynamic monitoring, providing real-time insights and facilitating proactive decision-making.

\section{Application Layer: Transforming Insights into Business Impact}

The system we built bridges the gap between technical data pipelines and actionable business intelligence by providing an intuitive interface designed for diverse users in the infrastructure and industrial sectors. This section highlights the key functionalities of the application layer, illustrating its ability to save time, enhance decision-making, and unlock new analytical possibilities. Here are only a select few examples of fundamental business process workflows how our system helps advance industrial operational efficiency. 

\subsection{Dynamic Filtering for Instant Insights}
\begin{figure}[ht]
    \centering
    \includegraphics[width=\linewidth]{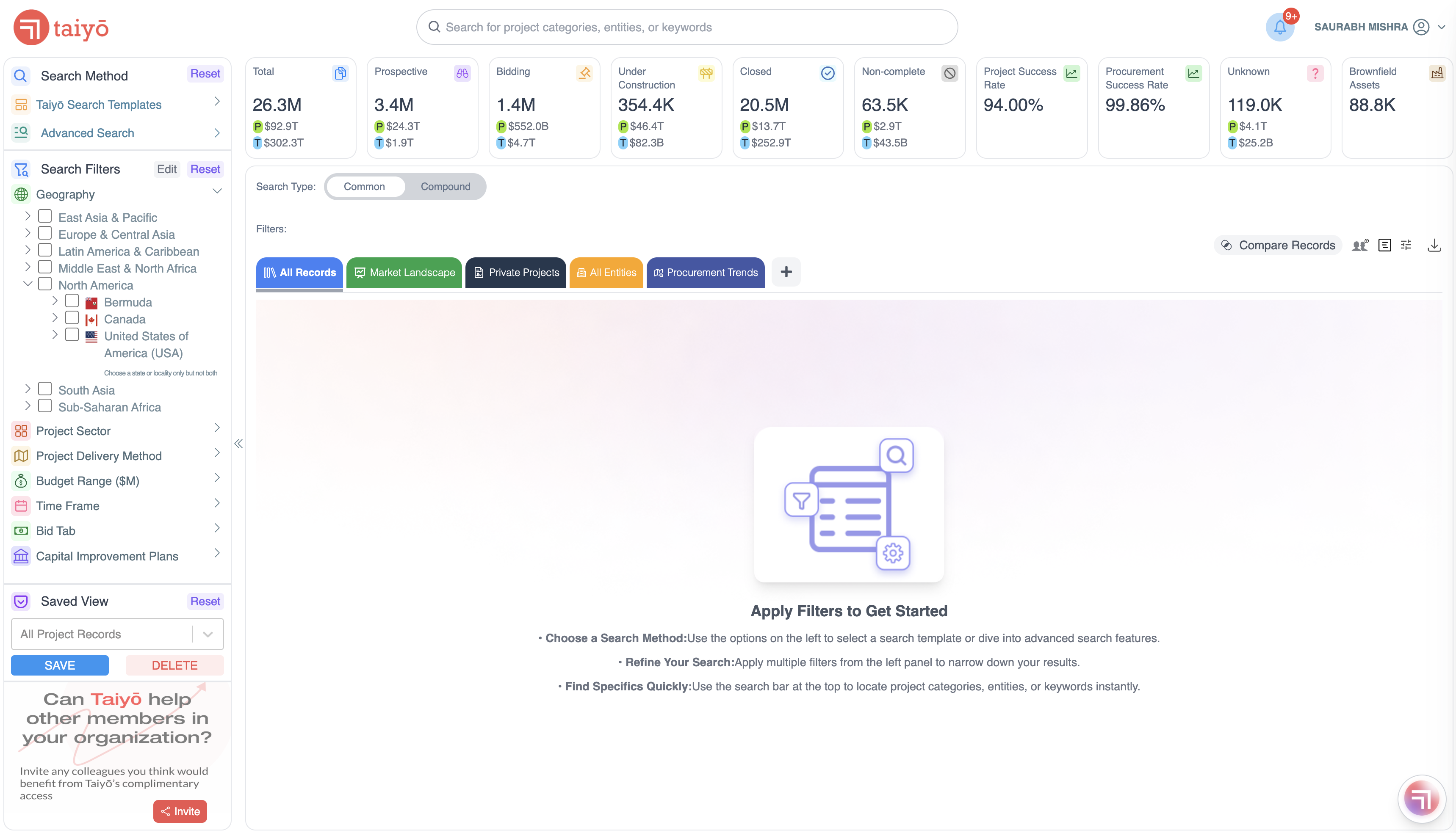}
    \caption{Dynamic filters on the left-hand and top navigation panels, enabling precise queries and instant updates.}
    \label{fig:filters_ui}
\end{figure}

The filtering capabilities, illustrated in Figure~\ref{fig:filters_ui}, are central to its usability. The left-hand navigation panel and the top filtering bar allow users to refine their queries dynamically. Filters can be applied across project sectors (e.g., solar, infrastructure, renewable energy), geographic regions (e.g., California, South Asia), and funding timelines. For example, a user interested in renewable energy projects in South Asia can quickly apply these filters, and the platform will update the data and visualizations in real-time. This streamlined functionality saves hours of manual data sorting and ensures decisions are based on precise, up-to-date information.

\subsection{Streamlined Business Development for EPCs and Large Construction Firms}
\begin{figure}[ht]
    \centering
    \includegraphics[width=\linewidth]{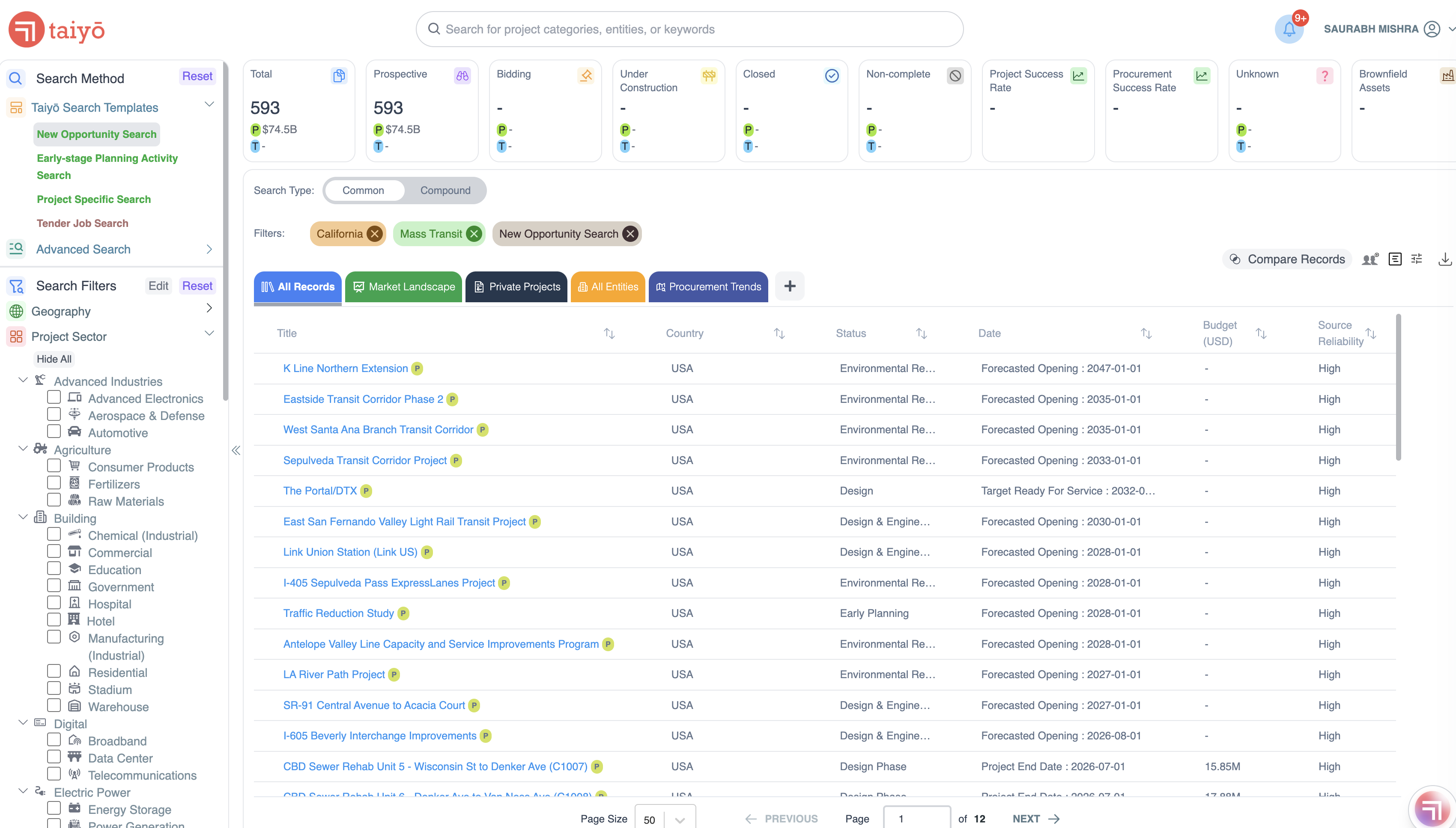}
    \caption{Business development workflow for construction firms, consolidating project pipelines across regions.}
    \label{fig:bd_workflow}
\end{figure}

The business development workflow for construction firms is often hindered by the time-intensive process of collecting, analyzing, and comparing pipelines across different regions and project types. The platform transforms this process by consolidating projects into a unified pipeline, as shown in Figure~\ref{fig:bd_workflow}. For instance, a firm interested in road construction in California, mass transit in New York, and rail projects in Europe can access all relevant data within minutes instead of spending weeks researching. This efficiency gain allows teams to focus on evaluating high-value opportunities, improving their profitability and strategic alignment with market trends.

\subsection{Optimized Sales and Marketing for Suppliers}
\begin{figure}[ht]
    \centering
    \includegraphics[width=\linewidth]{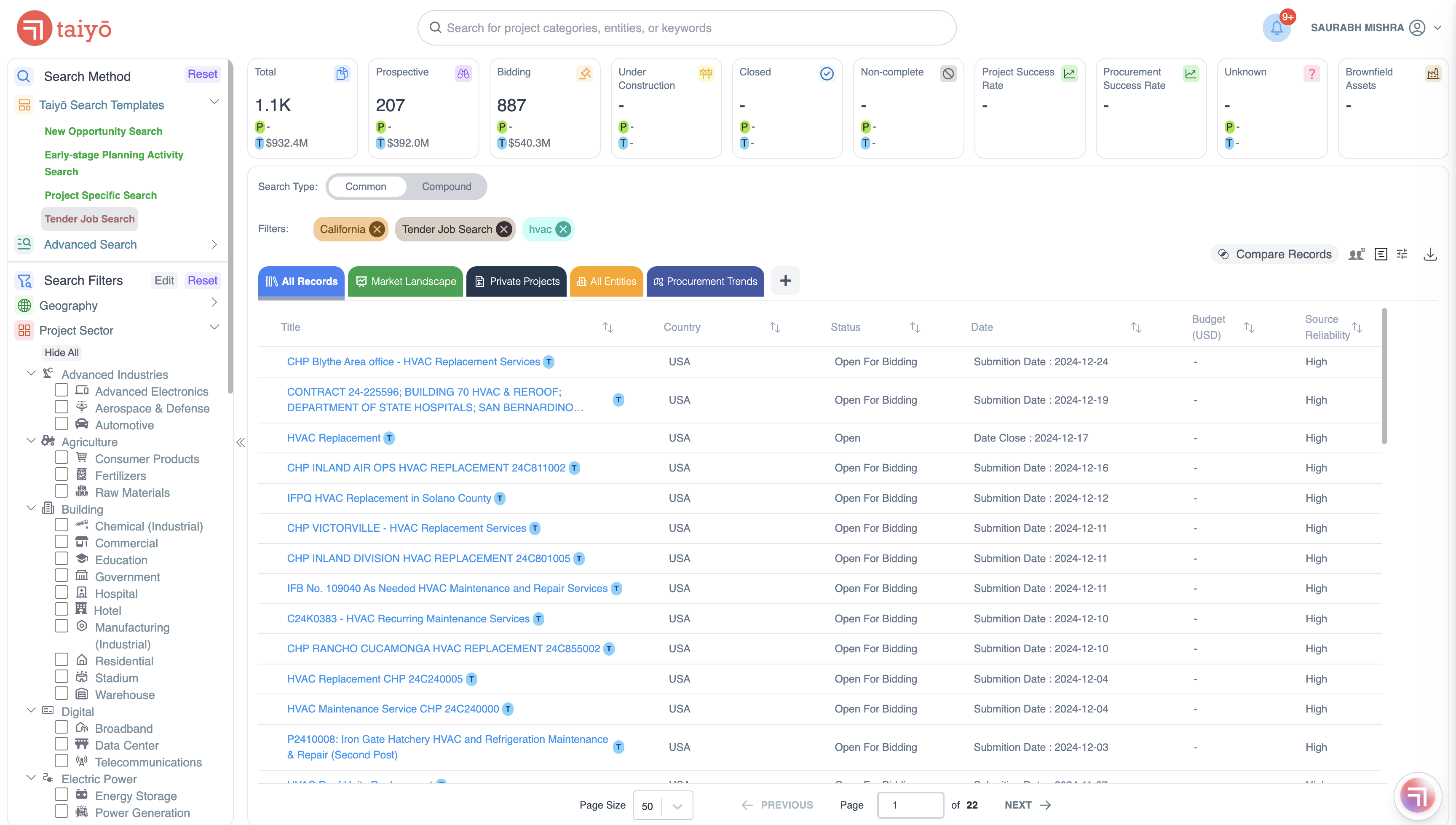}
    \caption{Supplier Insights module, designed to optimize sales and marketing workflows.}
    \label{fig:supplier_insights}
\end{figure}

Suppliers in the industrial and construction sectors often face challenges in staying updated on procurement opportunities and market trends. The platform simplifies this process by aggregating data from various procurement systems into a single platform, as seen in Figure~\ref{fig:supplier_insights}. Consider an HVAC supplier in California seeking to understand global procurement opportunities. Suppliers can analyze public procurement trends, identify competitive activities, and pinpoint new market opportunities, reducing research time by 70\% and enhancing their ability to act strategically.

\subsection{Sector-Specific Market Landscape for Strategic Decision-Making}
\begin{figure*}[ht]
    \centering
    \includegraphics[width=0.4\linewidth]{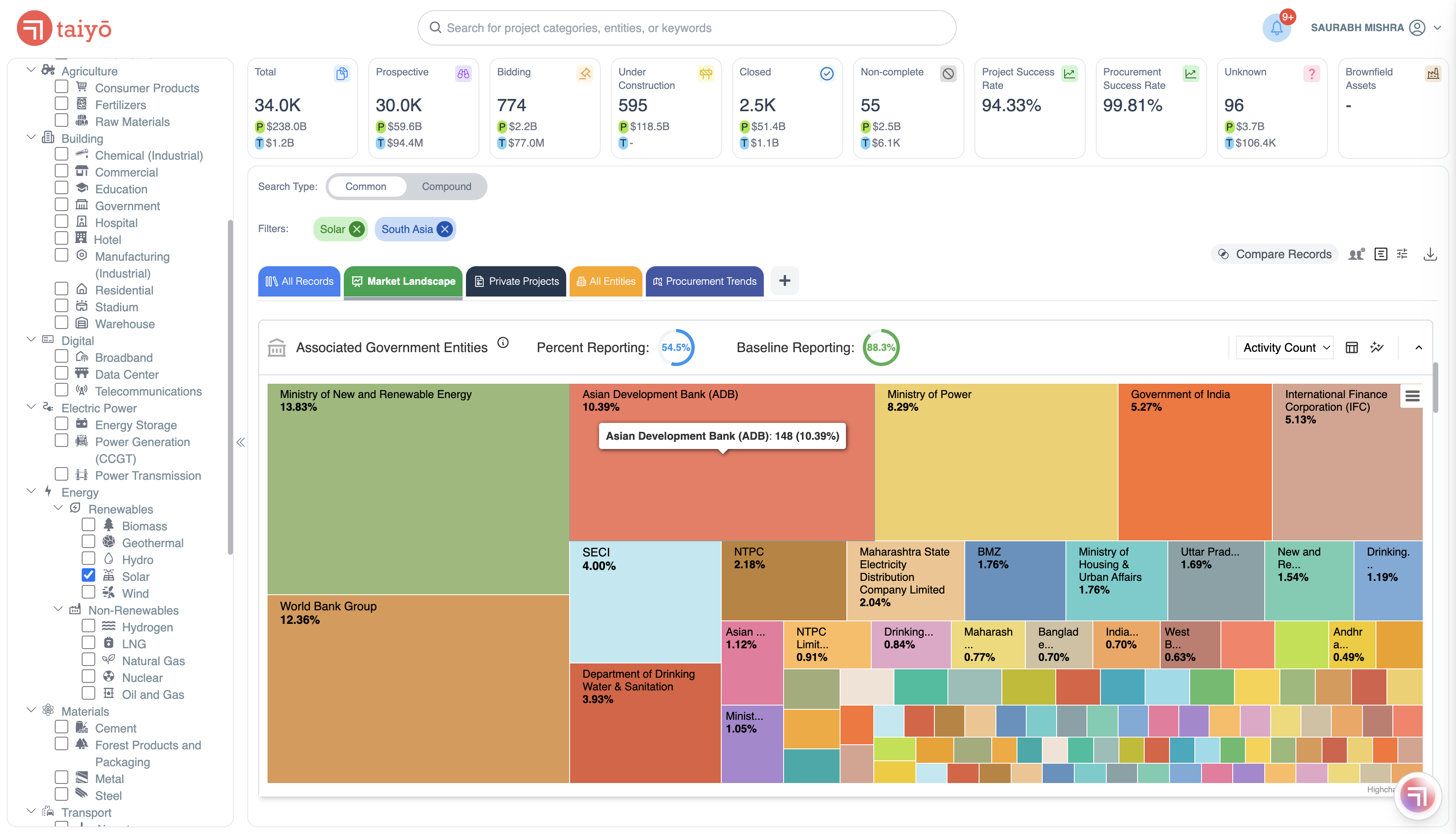}
    \vspace{-0.3cm} 
    \includegraphics[width=0.4\linewidth]{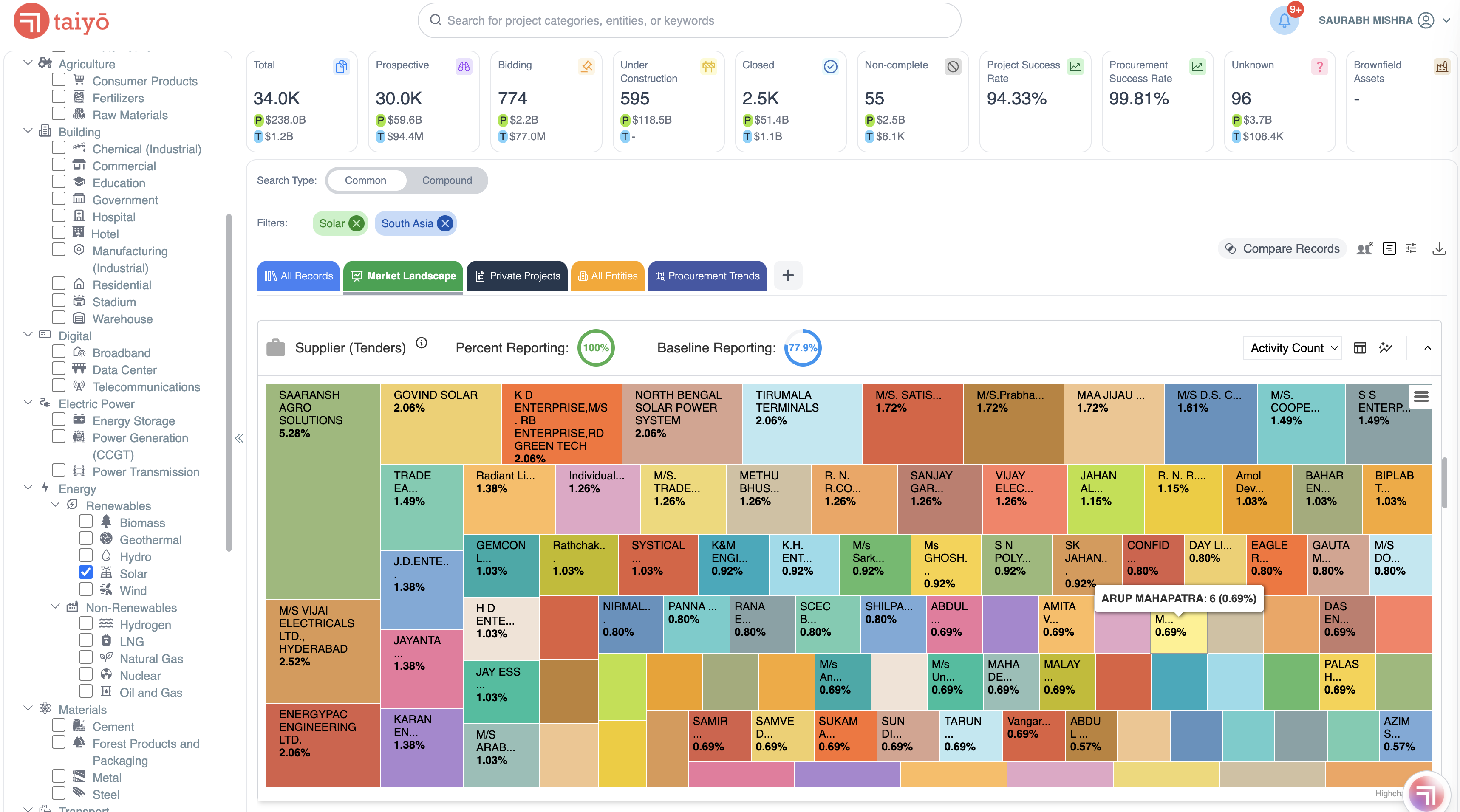}
    \vspace{-0.3cm}
    \includegraphics[width=0.5\linewidth]{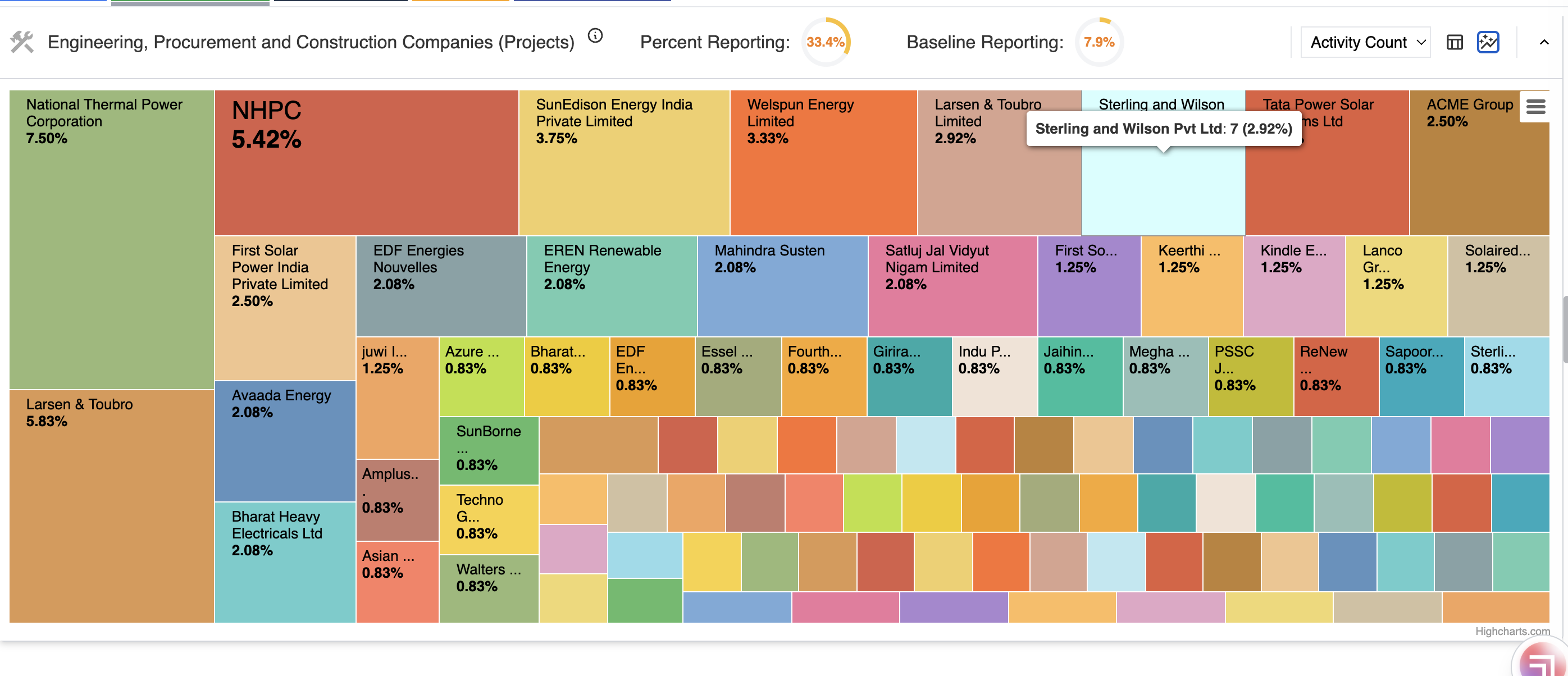}
    \vspace{-0.3cm}
    \includegraphics[width=0.5\linewidth]{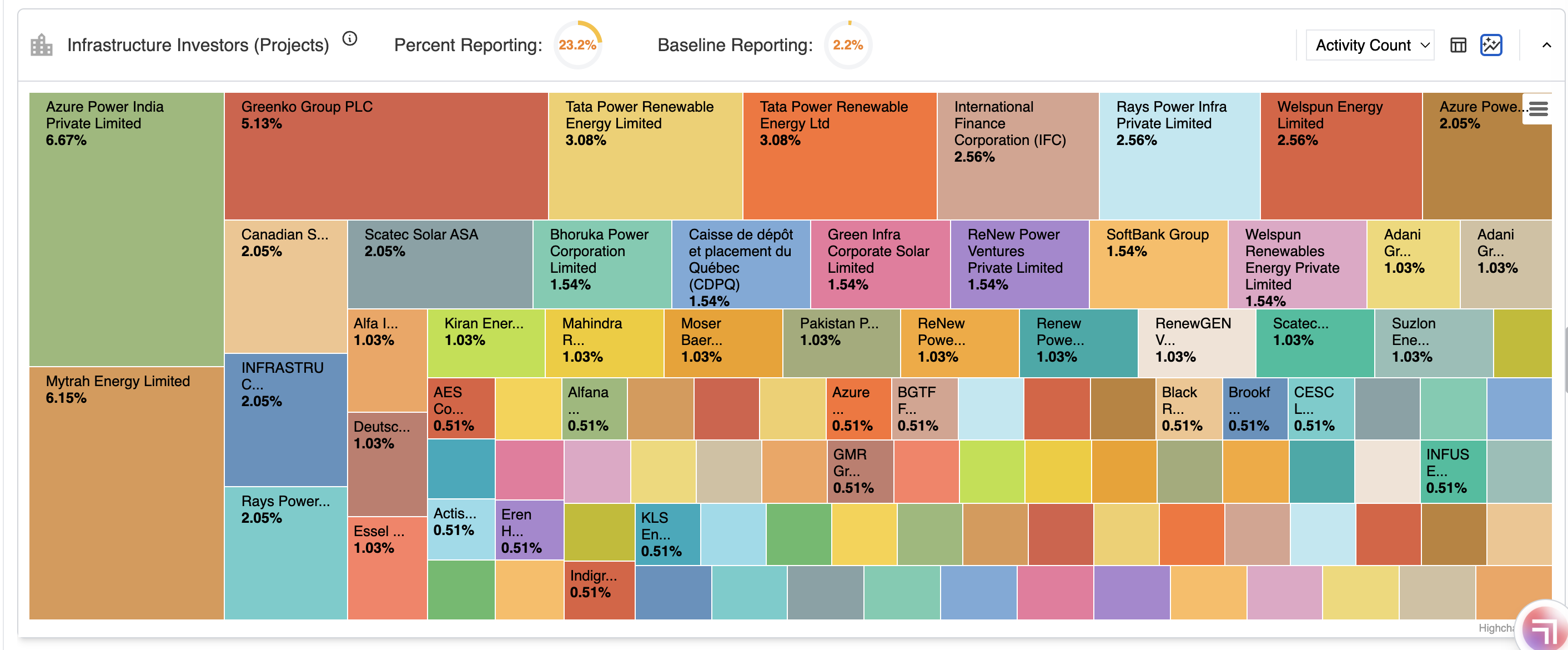}
    \includegraphics[width=0.5\linewidth]{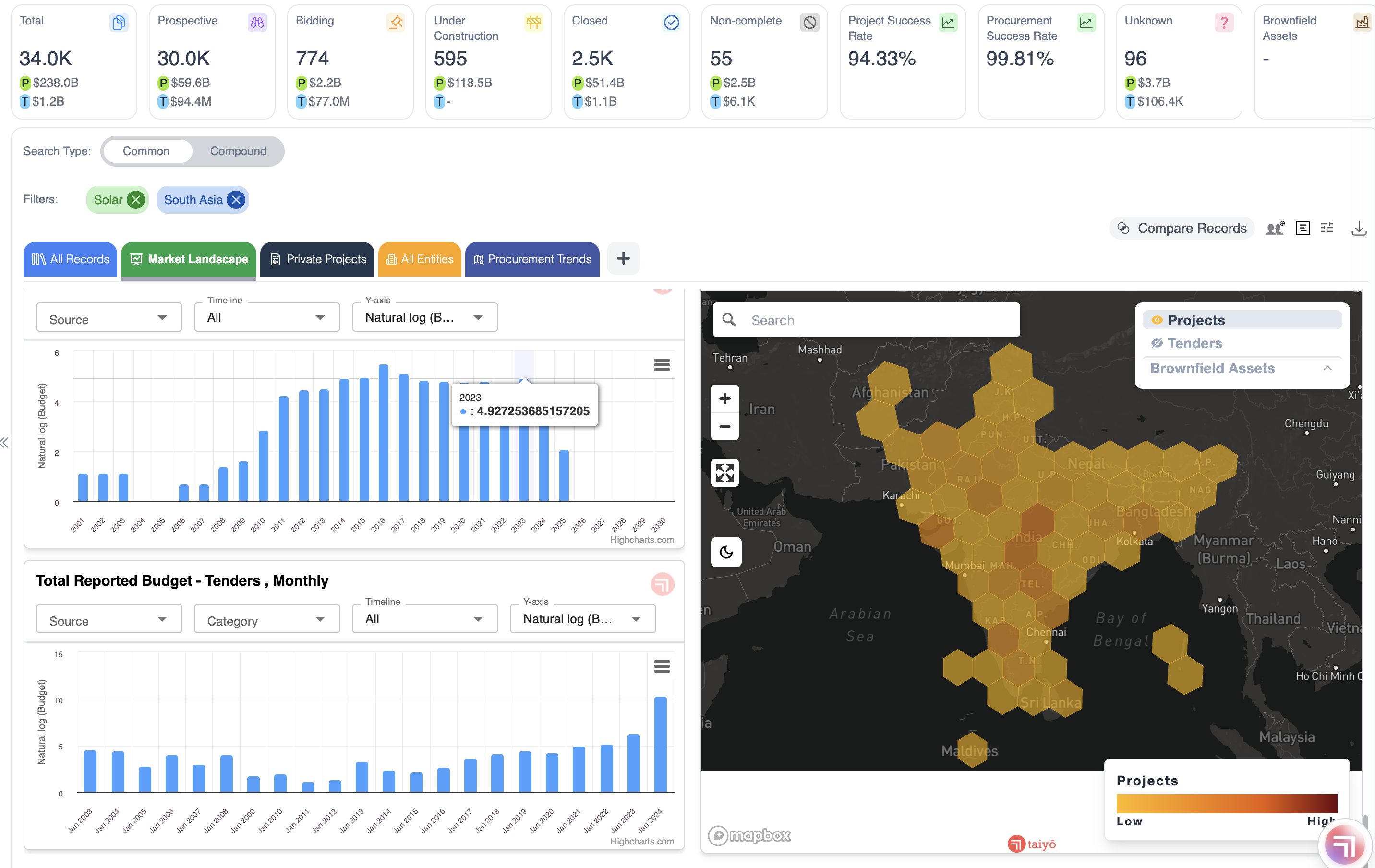}
    \caption{Visualizing Market Landscape for a given sector or region segmented by various entities group by project or procurement data sources}
    \label{fig:combined_images}
\end{figure*}

Figure~\ref{fig:combined_images} provides a comprehensive view of the Market Landscape module, which visualizes sector-specific data through tools such as heatmaps and treemaps. This module allows users to analyze top-performing entities, geographic activity, and sector-specific trends. For example, investors exploring the renewables sector in South Asia can quickly identify key players and emerging markets, empowering data-driven investment decisions. By consolidating fragmented market information, our approach not only saves time but also enhances the depth and accuracy of strategic planning.

\subsection{Generative AI for Systematic Learning and Real-Time Updates}
\begin{figure*}[ht]
    \centering
    \includegraphics[width=1\textwidth]{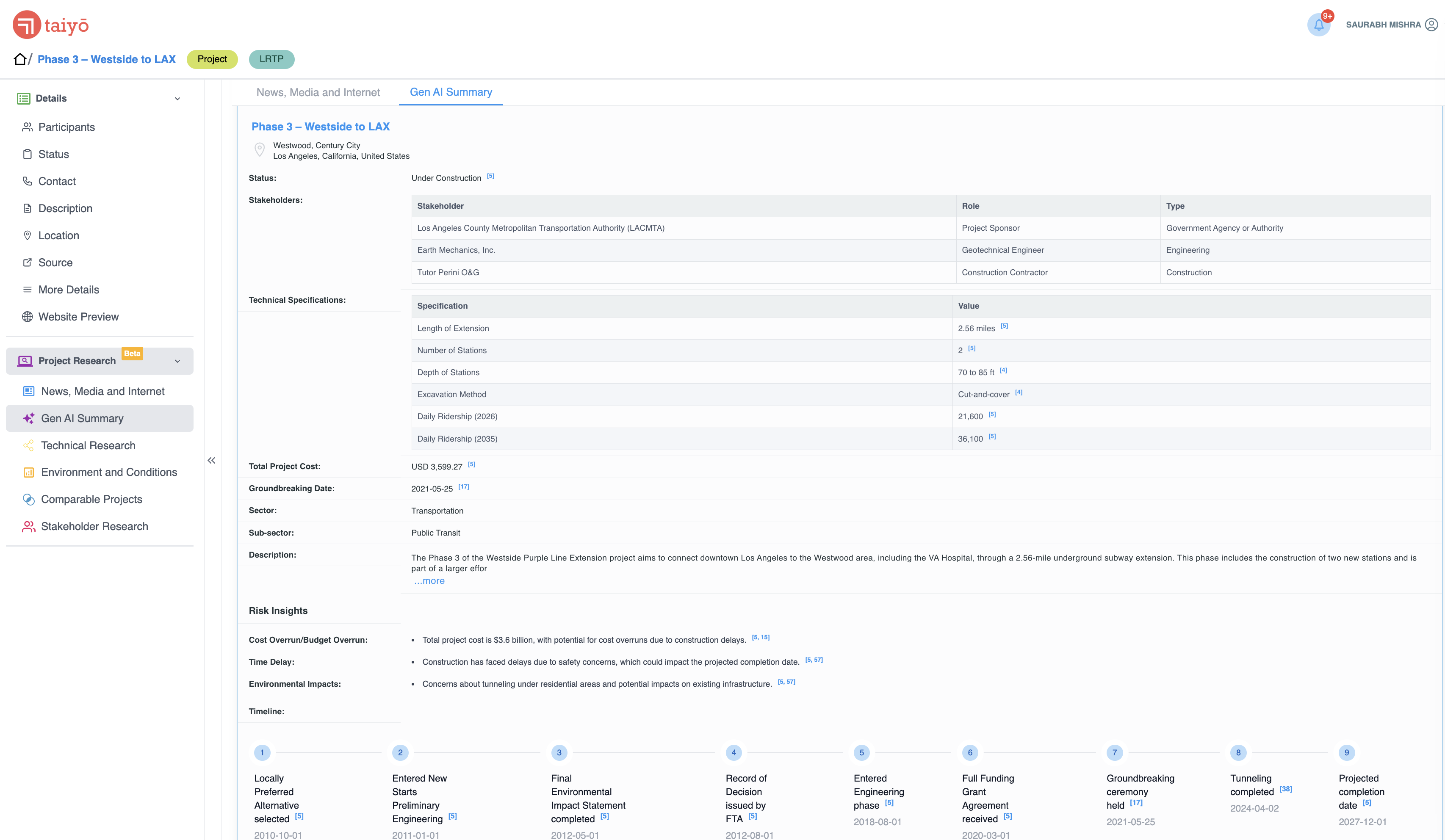}
    \caption{Project Summary and Tracker GenAI Agent Example, incorporating latest official project record and latest news and web information to structure a project summary}
    \label{fig:figlast}
\end{figure*}

The Generative AI capabilities automate the process of staying updated with the latest project developments. As shown in Figure~\ref{fig:figlast}, this module leverages AI agents to track real-time changes, such as project timelines, funding allocations, or regulatory updates. These insights are then integrated into users' workflows, ensuring decisions are always based on current data. For example, a construction firm can use this feature to receive alerts on shifting project scopes, enabling proactive adjustments to their plans. Generative AI adds a layer of automation and intelligence to the platform, complementing its other functionalities.

The platform not only streamlines existing workflows but also opens up new avenues for analytical work in the industrial and infrastructure sectors. The platform’s ability to aggregate, filter, and visualize data at scale makes previously impossible analyses feasible. Users can explore trends, benchmark performance, and identify emerging opportunities with unprecedented ease. While this section focuses on a few key features, the broader potential of our system its ability to transform data into actionable insights, empowering firms to operate more efficiently and strategically.

\section{Conclusion} \label{conclusion} 
Our AI-driven Data Mesh and Knowledge Graph platform represent an advancement in data management and governance, significantly enhancing decision-making processes in the infrastructure construction and public procurement industry. By integrating AI-driven automation with human expertise and rigorous data standards, we present a comprehensive framework designed to overcome key challenges faced by the industry, such as fragmented data sources and the absence of standardized practices.

This work offers a robust solution through a centralized AI system capable of managing and enriching both proprietary and web-streamed data in real time. Leveraging cutting-edge architectural elements, including a Service Mesh and Knowledge Graph, the platform ensures data accessibility, accuracy, and contextual enrichment. Such features facilitate informed, strategic decision-making, empowering industry practitioners to achieve higher productivity and efficiency. Our system addresses critical gaps by enabling real-time, quality-assured information enrichment and setting a new standard for data governance and transparency.

The platform aligns with global data standards, such as the Open Contracting Data Standard (OCDS), to promote openness and transparency in public procurement. By centralizing procurement information and standardizing workflows for red flag detection and supplier evaluation, our system aids governments in making well-informed decisions about funding recipients and hiring. We further streamline supplier tracking, offering tools for assessing financial stability, specialization, and SME or minority-owned status, which are crucial for equitable and efficient procurement.

However, the complexity of representing knowledge within large, interconnected infrastructure systems remains a significant challenge. Issues such as abstract boundary definitions in megaprojects and their sub-components complicate the knowledge representation process. Additionally, uncertainty quantification is essential to accelerate both human learning and the adaptive capabilities of AI models, requiring ongoing research and development.

Looking ahead, our work lays the foundation for expanding into more specialized AI sub-agents tailored to distinct analytical tasks, such as market analysis and supply chain optimization. We envision extending our platform’s capabilities to support business development, marketing, and sales workflows, transforming industrial sectors and enabling global competitiveness. Furthermore, future research will focus on enhancing the supply-side component of our platform, with deeper integration of inventory and procurement data. This expansion will facilitate better purchasing decisions for industrial equipment and services, streamlining supply chain operations and maximizing efficiency.

We also aim to extend our Knowledge Graph to cover procurement volumes, product types, and evolving standards, enhancing the system’s versatility and applicability in other sectors. Such efforts will provide a more competitive and transparent marketplace, fostering cross-border industrial growth and collaboration.

The potential impact of our platform extends beyond the infrastructure and construction sectors. By providing foundational information and intelligence, our system sets the stage for powerful transformations in any analog industry seeking to transition to digital workflows. The integration of comprehensive data governance, enhanced transparency, and structured knowledge has significant implications for tackling climate change, making smarter procurement choices, and fostering sustainable economic growth.

\bibliographystyle{unsrt}
\bibliography{references}

\newpage 

\newpage 

\end{document}